\documentclass[10pt,journal,compsoc]{IEEEtran}

  \usepackage{cite}

\usepackage{mathptmx}
\usepackage{graphicx}
\usepackage{times}

\graphicspath{{images/}}

\usepackage[T1]{fontenc}
\usepackage[english]{babel}
\usepackage{graphicx}
\usepackage{subfigure}
\usepackage{amsmath}
\usepackage{amssymb}
\usepackage{color}
\usepackage{times}
\usepackage{epsfig}
\usepackage{caption}
\usepackage{diagbox}
\usepackage{stmaryrd}
\usepackage{algorithm}
\usepackage[noend]{algpseudocode}

\newcommand{\subparagraph}{}
\usepackage{titlesec}
\titleformat{\paragraph}
{\normalfont\normalsize\bfseries}{\theparagraph}{1em}{}
\titlespacing*{\paragraph}
{0pt}{3.25ex plus 1ex minus .2ex}{1.5ex plus .2ex}

\newcommand{\etal}{\textit{et al. }}

\usepackage{diagbox}
\usepackage{subfigure}

\newcommand{\mbf}[1]
{
	\mathbf{#1}
}

\newcommand{\vect}[1]{\mathbf{#1}}

\ifCLASSINFOpdf
\else
\fi

\begin{document}

\title{A Multiple-View Geometric Model for Specularity Prediction
on General Surfaces}

\author{Alexandre~Morgand,
        Mohamed Tamaazousti
        and~Adrien~Bartoli
\IEEEcompsocitemizethanks{\IEEEcompsocthanksitem  A. Morgand is with SLAMcore ltd, London, UK
\protect\\
E-mail: morgand.alexandre@gmail.com
\IEEEcompsocthanksitem M. Tamaazousti is with Universit\'e Paris Saclay, CEA, LIST, Gif-sur-Yvette, France. \protect\\
E-mail: mohamed.tamaazousti@cea.fr
\IEEEcompsocthanksitem A. Bartoli is with IP-UMR 6602 - CNRS/UCA/CHU, Clermont-Ferrand, France. \protect\\
E-mail: adrien.bartoli@gmail.com}
\thanks{}}

\markboth{IEEE TRANSACTIONS ON VISUALIZATION AND COMPUTER GRAPHICS 2022}%
{Morgand \MakeLowercase{\textit{et al.}}:A Multiple-View Geometric Model for Specularity Prediction on Non-Uniformly Curved Surfaces}

\IEEEtitleabstractindextext{%
\begin{abstract}
Specularity prediction is essential to many computer vision applications, giving important visual cues usable in Augmented Reality (AR), Simultaneous Localisation and Mapping (SLAM), 3D reconstruction and material modeling. However, it is a challenging task requiring numerous information from the scene including the camera pose,  the geometry of the scene, the light sources and the material properties. Our previous work addressed this task by creating an explicit model using an ellipsoid whose projection fits the specularity image contours for a given camera pose. These ellipsoid-based approaches belong to a family of models called JOint-LIght MAterial Specularity (JOLIMAS), which we have gradually improved by removing assumptions on the scene geometry. However, our most recent approach is still limited to uniformly curved surfaces.
This paper generalises JOLIMAS to any surface geometry while improving the quality of specularity prediction, without sacrificing computation performances. The proposed method establishes a link between surface curvature and specularity shape in order to lift the geometric assumptions made in previous work. Contrary to previous work, our new model is built from a physics-based local illumination model namely Torrance-Sparrow, providing an improved reconstruction.
Specularity prediction using our new model is tested against the most recent JOLIMAS version on both synthetic and real sequences with objects of various general shapes. Our method outperforms previous approaches in specularity prediction, including the real-time setup, as shown in the supplementary videos.

\end{abstract}

}

\maketitle

\IEEEdisplaynontitleabstractindextext

%
\IEEEpeerreviewmaketitle

\IEEEraisesectionheading{\section{Introduction}\label{sec:introduction}}

\IEEEPARstart{L}{ight} is of tremendous interest in many fields of science, including physics and computer vision. In computer graphics, visual effects and AR, illumination plays a crucial role in rendering an object realistically. However in AR, mixing virtual objects and real scene elements is particularly challenging to achieve because the lighting conditions of the virtual scene have to match the real ones while maintaining real-time rendering performances. 
Among the important lighting elements to render, the specular reflections, also called specular highlights or specularities, are given as the most important elements to render to improve the realism of virtually inserted objects\cite{artusi2011survey, delpozo2007detecting, kim2013specular, klinker1990physical}. In human perception, according to Black \etal \cite{blake1990does}, specularities are essential elements to distinguish shapes and the motion of shiny objects.
Isolating a specularity in an image is challenging because its appearance depends on numerous elements in the scene, including the camera pose, the geometry of the scene, the light sources, the roughness and material properties of the surface. This dependency is important because in an AR context, rendering specularities correctly emphasises the perception of an object material and its geometry, should the specularity motion be coherent. However, accessing all these elements is still an open problem since knowing the lighting conditions requires one to compute many parameters. The specularity's motion may also vary through the video stream due to the varying surface geometry (flexible surfaces), different types of materials present in the scene or light sources that could be switched on and off. To sum up, the task of specularity prediction is challenging but offers a strong potential realism in AR applications, motivating its study in the current literature\cite{karsch2011rendering, lombardi2012reflectance, buteau2015poster, jachnik2012realtime, meilland20133d}.

Generally speaking, the existing methods fall in three categories: (1) light source modelling methods, including global illumination, which reconstruct the whole lighting context of the scene, and local illumination models, computing the 3D position of the illumination and specularity prediction; (2) deep learning methods using synthetic and real databases of images for novel viewpoint generation through latent space interpolation and (3) multiple-view reconstruction approaches. Aside of these categories, methods \cite{park2020seeing, richter2016instant, park2018surface} do not model specularities explicitly but include them in a process of lighting condition reconstruction. They require one to compute numerous parameters on the materials and light sources, and cannot predict the specularity on new viewpoints. 

Category (3) is the most recent, which we have proposed. We have shown in \cite{morgand2016empirical, morgand2017geometric, morgand2017multiple, morgand2017amultiple} that a specularity on a planar surface can be well approximated by an ellipse under a light bulb or a fluorescent lamp illumination found commonly indoors. We have proposed a family of models called JOint LIght-MAterial Specularity (JOLIMAS) to abstract the light-matter interaction and treat the problem geometrically. This family of models uses a fixed 3D ellipsoid whose projection predicts the specularity's shape in new viewpoints. The 3D ellipsoid reconstruction is achieved from at least three viewpoints. However, the existing approaches are restricted to planar or slightly curved surfaces, which is not ideal in a real environment including curved surfaces.
JOLIMAS requires camera pose and scene structure to be available. In practice, owing to the recent advances in visual SLAM, including numerous methods being publicly available, providing real-time camera localisation \cite{engel2017direct, campos2020orb, schneider2018maplab, tamaazousti2016constrained} and 3D reconstruction \cite{newcombe2011kinectfusion, whelan2015elasticfusion}, obtaining quality camera pose and 3D geometry of the scene in real-time as inputs for specularity prediction is now possible.

We describe a twofold contribution in geometric specularity modelling. First, we present a generalisation of the specularity prediction model JOLIMAS to {\em any} surface geometry. This works by transforming the specularity contours according to the local curvature of the surface. The idea behind this transformation is to model how these contours would be if the surface were deformed to a plane using the Torrance-Sparrow local reflectance model\cite{torrance1967theory}. This represents the first attempt to introduce a physics-based model to the JOLIMAS modelling approaches. Then, we reconstruct a canonical representation of the JOLIMAS model as an ellipsoid from the transformed contours. This ellipsoid will remain constant for any surface deformation associated to the specularity of interest. Secondly, we propose an inverse transformation of these contours when performing specularity prediction. This prediction is obtained by projecting the ellipsoid from our canonical model to the image and transforming the contours to fit the current local curvature. This process runs in real-time for objects with known geometry in the scene and with  camera pose as inputs. 

First, we present the related work in specularity prediction in section \ref{sec:related_work} to position our method within the state-of-the-art. We then state the formal specularity prediction problem and give background in section \ref{sec:background} to propose the canonical dual JOLIMAS approach in section \ref{sec:canonical}.
To assess the efficiency of our method, we conduct two quantitative experiments on synthetic data and show qualitative results on real sequences for objects of various curvature in section \ref{sec:results}. 

\section{Related Work}
\label{sec:related_work}

Modelling lighting effects like shadows and specularities is a complex problem. This is because they depend on interdependent elements such as surface material, geometry and roughness. It is thus relevant to focus first on the specularity prediction task, which is less strongly dependent and a logical first step for light conditions understanding. It is also interesting to consider specularities due to their importance in computer vision. By adding variations to the image intensity, numerous algorithms are affected by specularities in the image such as segmentation, reconstruction and camera localisation methods. However, specularities provide relevant information on the light sources, the scene geometry and materials.
Due to specularity prediction being a fairly recent research topic, the state-of-the-art is yet limited.
We describe the three major specularity prediction approaches: light sources modelling methods using global and local illumination models, deep learning methods and geometric modelling methods.

\subsection{Light Source Modelling Methods}
\paragraph*{Global illumination models}

Methods in this category favour the rendering quality by using the rendering equation \cite{immel1986radiosity, kajiya1986rendering}. This equation describes the total amount of light emitted from a point along a particular viewing direction, given a function for the incoming light and a Bidirectional Reflectance Distribution Function (BRDF). These approaches do not generally compute the physical attributes of the light sources. For instance, \cite{jachnik2012realtime} captures a 4D light field over a small planar specular surface. By reconstructing the diffuse and specular components, it achieves convincing rendering. However, it is unable to predict the specular component for new viewpoints unseen during initial reconstruction. Moreover, light sources with changing states (turned on and off) are not handled. Recently, \cite{richter2016instant} extended \cite{jachnik2012realtime} by adding material segmentation for complex surfaces reconstructed using an RGB-D sensor, but shared the same limitations as \cite{jachnik2012realtime}. As a consequence, \cite{jachnik2012realtime, richter2016instant} and similar approaches such as \cite{meilland20133d} cannot predict specular reflections for new viewpoints and changing lights.

Despite the quality of the results, these methods lack flexibility. Indeed, they represent an extended light source such as a fluorescent lamp by several point light sources. Thus, these methods need to compute the lighting intensity of different point light sources instead of treating them as a unique light source. Moreover, dynamic lighting cannot be used and materials for the predicted specularity are not computed. As a consequence, these methods are not suited to specularity prediction for unknown viewpoints. Also, due to a long initialisation process, these methods require numerous images and processing power which is not adapted to the real-time context of AR.

\paragraph*{Local illumination models}
In parallel, other works have been proposed on primary light source reconstruction. Ideally, each light source should be associated to a geometry (position and shape), a colour and an intensity value in order to match the scene\cite{lindsay2014automatic, lin2017color} realistically. Although numerous light source models exist in computer graphics, the models used in computer vision are generally divided in two categories: directional light sources and point light sources. For external scenes, a directional light source feels more natural to represent the sun but could also be useful for indoors scenes (ceiling light sources for instance). \cite{lagger2006using} describes a method to compute directional light sources from a fixed viewpoint. The application is limited because specularities are strongly dependent on the viewpoint and the light sources need to be estimated for each pose. Neither the shape and position of the light source, nor the object material are taken into account, making this method unable to predict specularities explicitly and accurately. Moreover, extended light sources such as fluorescent lamps cannot be modeled correctly by point light sources, limiting the method's applicability.

Related approaches \cite{boom2013point, buteau2015poster, einabadi2015discrete, kanbara2004real, wong2008recovering} suffer from similar issues and limitations. For example, in light source estimation, \cite{boom2013point} computes a primary light source under the Lambertian assumption and using an RGB-D camera. From the diffuse term and by comparing a synthetic rendering of the scene with the current image, a unique point light source is estimated. However, this method is not adapted to real-time applications with multiple light sources that could be point light sources such as a desktop lamp or a light bulb, or extended sources like a fluorescent lamp. By computing only one light source without knowledge of the shape and materials of the surfaces of the scene, it is challenging to perform realistic and coherent specularity prediction.

Specularity prediction from a parametric light source reconstruction requires the computation of numerous complex and often ambiguous light and material parameters. For instance, the joint estimation of material reflectance and light source intensity is an ill-posed problem, as stated in \cite{lombardi2012reflectance}. In practice, a physics-based approach like the aforementioned ones cannot effectively predict a photometric phenomena as complex as a specularity.


\subsection{Deep Learning Methods}
The impact of deep learning methods in computer vision is growing rapidly, owing to the recent increase in processing power thanks to GPU's and CPU's, sometimes solely dedicated to machine learning.
The current access to huge labelled image datasets such as ImageNet\cite{deng2009imagenet}, COCO\cite{lin2014microsoft} and pretrained models such as VGG-19\cite{Simonyan15} and ResNet\cite{he2016deep} for feature extraction democratised the use of deep learning. The lighting modelling field is no exception to this trend. For deep global illumination, \cite{mandl2017learning} takes a 3D object with known albedos as inputs. Their Convolutional Neural Network (CNN) takes the coefficients of the spherical harmonics function computed as inputs in order to generalise the illumination after training. This method shows results from a single image but requires a training time of four hours per pose and the illumination is limited to the learnt object, which limits the range of applications of the method.
\cite{legendre2019deeplight} and more particularly \cite{park2020seeing} were an interesting continuation of \cite{jachnik2012realtime}, showing impressive results in terms of specularity prediction for new viewpoints by modelling highly specular objects using an RGB-D camera to reconstruct a light field. \cite{park2020seeing} was also able to model inter-reflections showing how far novel view synthesis could go using an adversarial approach. However, this method remains limited to the object of interest and does not provide strong insights on scene understanding in case the lighting conditions change or the camera viewpoint drastically changes.

Another interesting recent approach, perhaps more related to 3D inpainting, is \cite{aliev2019neural}. It generates novel viewpoints from RGB-D and point clouds data. Without any prior in terms of lighting modelling, this method interpolates what the lighting would be for a novel viewpoint. Even though this method is not directly aiming at realistic lighting conditions, it highlights the interpolation capabilities of deep learning based methods and shows how the recent methods in the state-of-the-art require heavy training and processing. This is not suitable for real-time application and lacks flexibility when the camera moves too far from the initial scene or if the scene is too different from the training database.  Additionally, very few image databases contain lighting information and more particularly annotated specularity images due to the difficulty to segment specularities in the images manually. To address this issue, synthetic databases are a decent solution to have a distinct specular-diffuse component separation but are sometimes not enough to generalise the network to real images. More recently, a new family of approaches called Neural Radiance Fields (NeRF) has greatly improved the quality of novel viewpoint rendering for non-Lambertian reflectance \cite{mildenhall2020nerf}. Even though this area of research is improving rapidly in terms of rendering speed and quality in the presence of slight object variations, NeRF methods require a long learning time and great computational power to make the rendering real-time. Moreover, these methods do not explicitly describe the light interactions but blindly output the RGB-D and radiance information for any given point.

\subsection{Geometric Modelling Methods}

\paragraph*{Principle}

\begin{figure}[!ht]
\subfigure[]
{
    \includegraphics[width=0.6\linewidth]{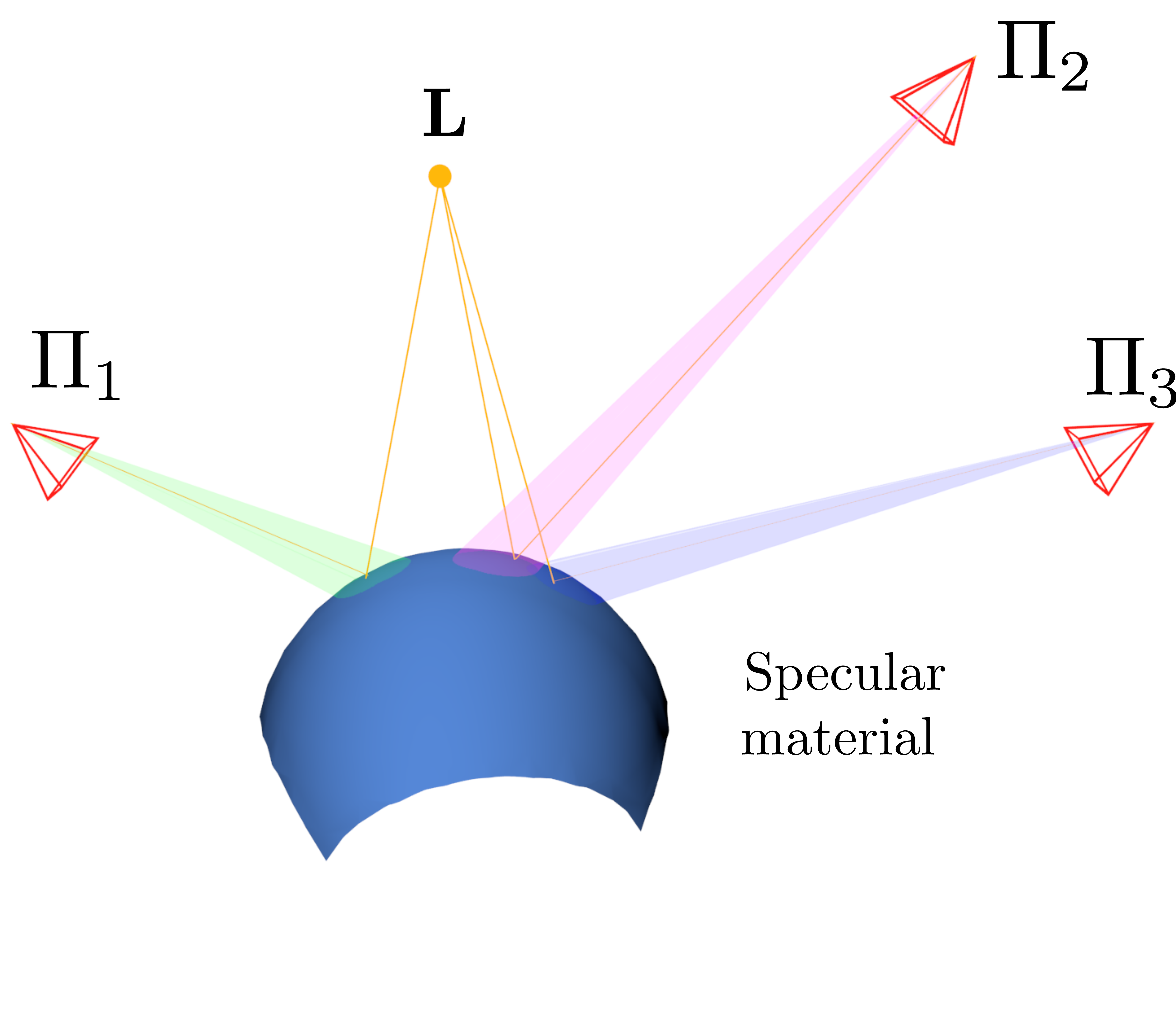}
}
\subfigure[]
{
    \includegraphics[width=0.3\linewidth]{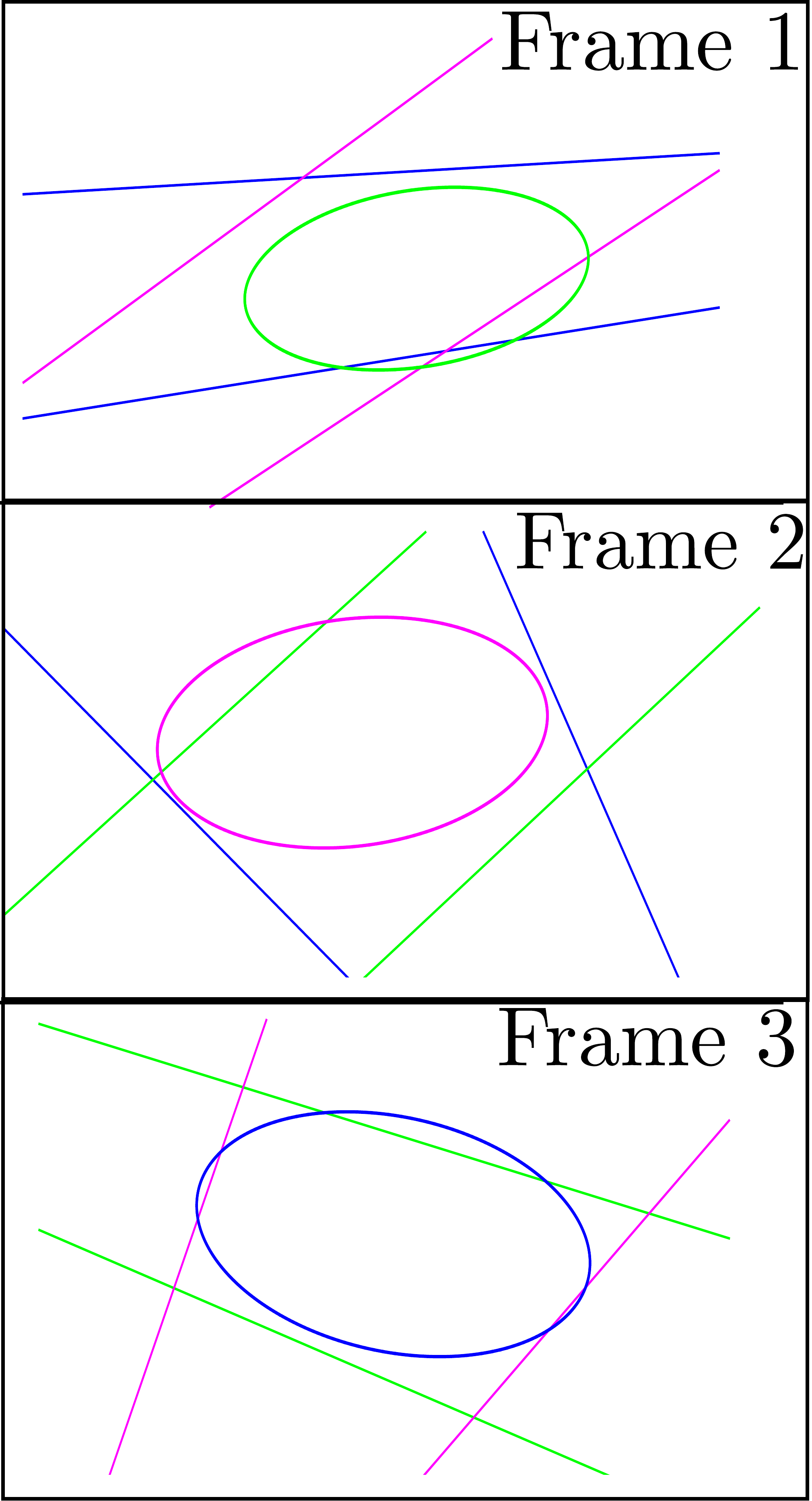}
    \label{fig:epipolar_lines_primal}
}
\caption{Epipolar geometry of the ellipses for three camera poses $\Pi_1$, $\Pi_2$ and $\Pi_3$  and a point light source $\mbf{L}$ using Primal JOLIMAS on a sphere. The scene is showed in (a) and the associated ellipses and epipolar lines in (b). The epipolar geometry is not respected. This results in an incorrect 3D ellipsoid reconstruction and incorrect specularity prediction. Figure extracted from \cite{morgand2017multiple}.}
\label{fig:epipolar_primal}
\end{figure}

Our previous work initiated a family of approaches called JOLIMAS as an original alternative and complement to other methods in specularity prediction.
It addresses several limitations by considering specularities as a geometric cue in the scene, as opposed to classic photometric approaches like BRDF estimation and inverse rendering.
The first version, {\em Primal JOLIMAS}\cite{morgand2017geometric}, showed that for planar glossy surfaces, a specularity created from a light bulb or fluorescent lamp has an elliptical shape. By reconstructing a fixed 3D ellipsoid from several image ellipses, this method predicts the specular shape for new viewpoints by simple perspective projection of the 3D ellipsoid. It abstracts light and material interactions, and can be used for retexturing. Specifically, primal JOLIMAS uses the following hypotheses:
\begin{enumerate}
    \item Specular reflections are elliptical on planar surfaces.
    \item A light source is generally associated with a single specularity on planar surfaces.
    \item There is a unique fixed 3D ellipsoid located `under' the planar surface whose perspective projection fits the specular shape in the image.
\end{enumerate}
Hence, Primal JOLIMAS fails to predict the specularity if the surface is non-planar, which drastically limits its applicability. For non-planar surfaces, the reconstructed 3D ellipsoid is not fixed in space because ellipses are not epipolar consistent, as illustrated in figure \ref{fig:epipolar_primal}. The problem is more complex because the reflected image of the light source (specularity) through a curved surface (mirror or specular surface) can be drastically distorted. To solve this issue, the distortion of the specularity should be included in the model. 

\begin{figure}[!ht]
\subfigure[]
{
    \includegraphics[width=0.6\linewidth]{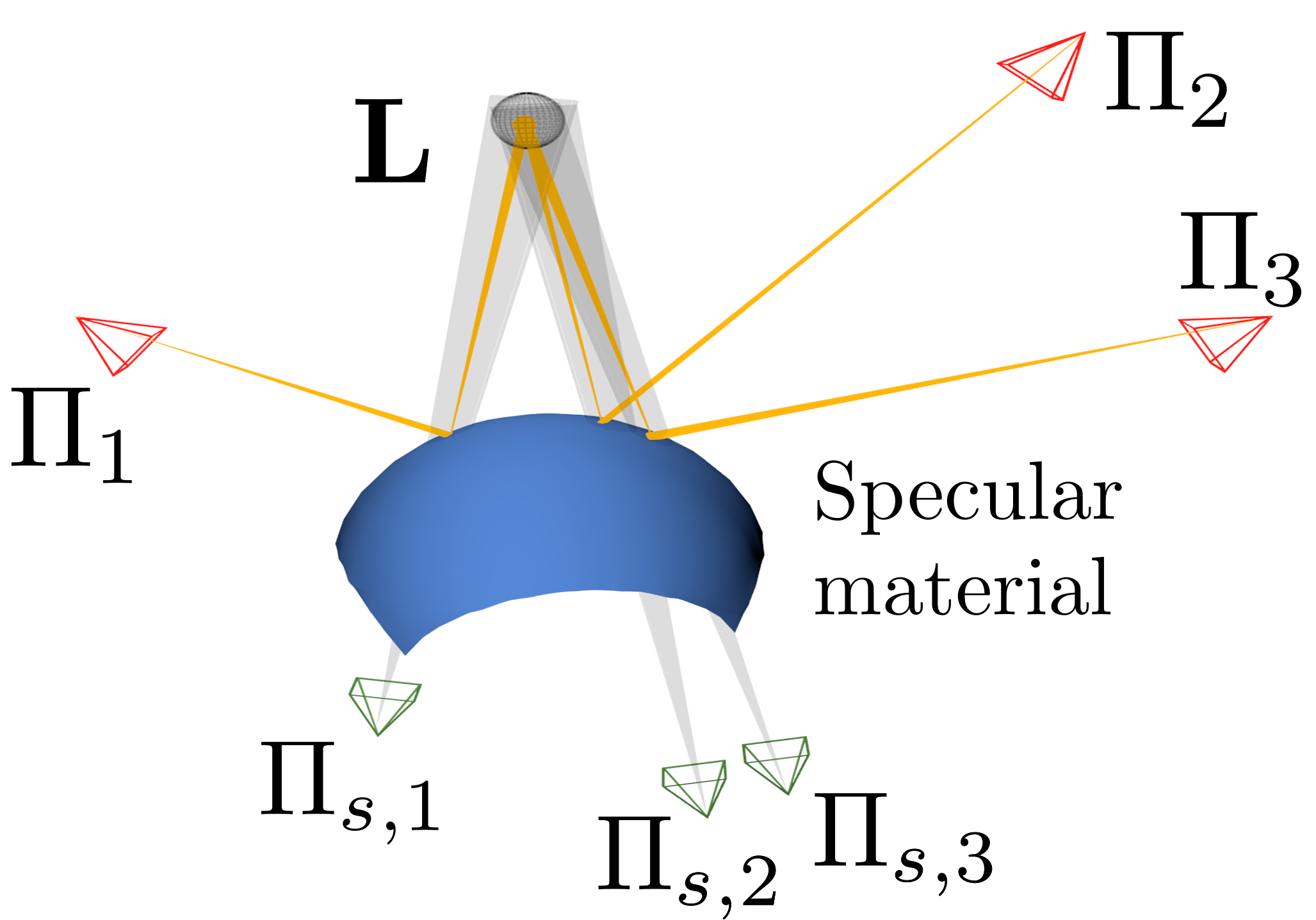}
}
\subfigure[]
{
    \includegraphics[width=0.3\linewidth]{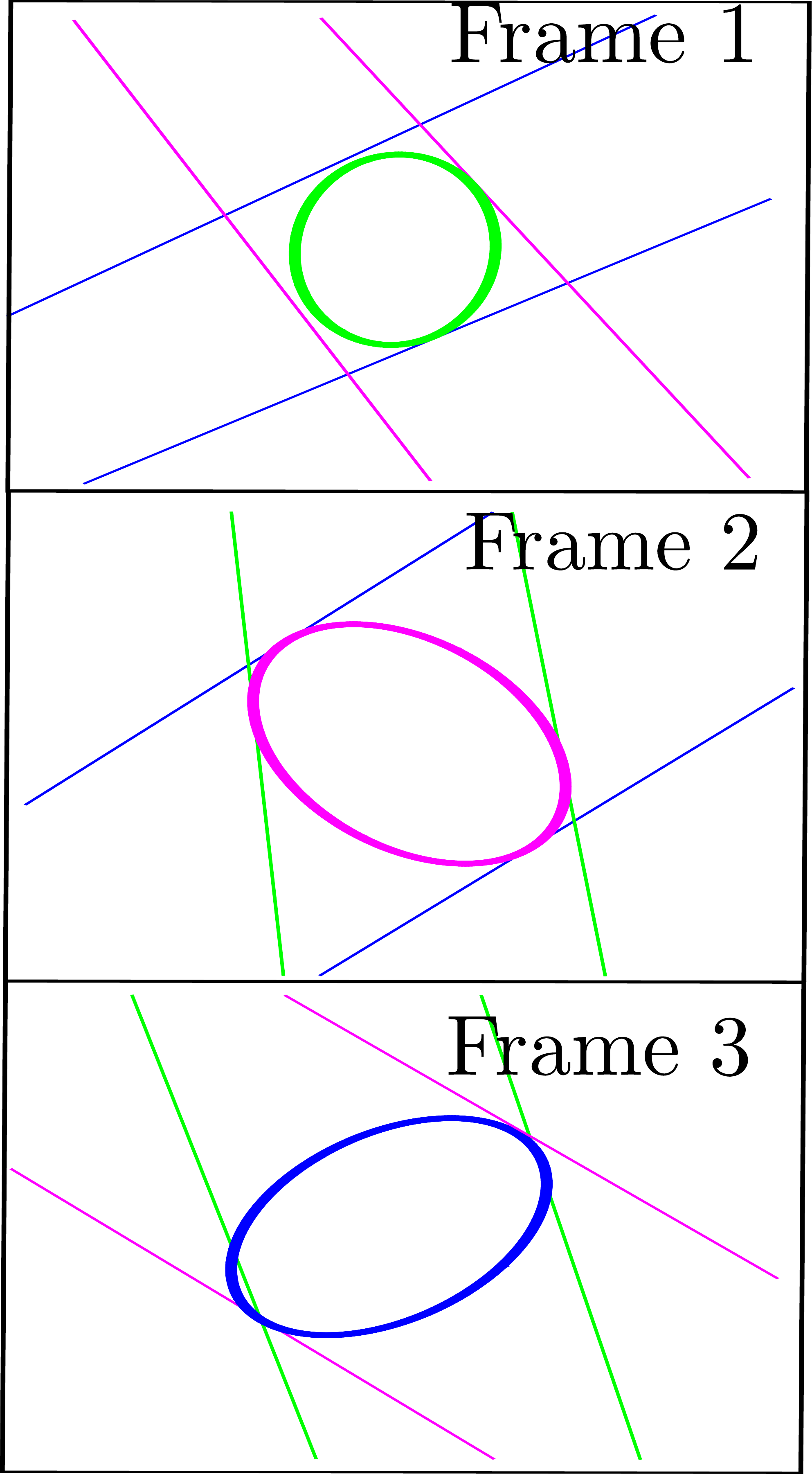}
}
\caption{Illustration of Dual JOLIMAS \cite{morgand2017multiple}, addressing the issue of specularity prediction on non-planar surfaces. In (a), the computation of virtual cameras $\Pi_{s,1}$, $\Pi_{s,2}$ and $\Pi_{s,3}$ from the real cameras $\Pi_1$, $\Pi_2$ and $\Pi_3$ is shown, allowing one to reconstruct an ellipsoid located near the light source. In (b), as opposed to figure \ref{fig:epipolar_lines_primal}, the epipolar lines fit the ellipse. Figure extracted from \cite{morgand2017multiple}.}
\label{fig:dual_jolimas}
\end{figure}

We proposed a second, extended version called {\em Dual JOLIMAS}\cite{morgand2017multiple},
 which uses virtual cameras computed from the camera poses reflected by the tangent plane at the brightest point of the specularity observed at the current pose. This virtual camera system was enough to reconstruct the 3D ellipsoid and predict specularities on non-planar surfaces with constant curvature, such as spheres and cylinders, as shown in figure \ref{fig:dual_jolimas}.

A high-level geometric representation of the light sources and more particularity specularities is beneficial for many applications. In the current state, the JOLIMAS approaches have a strong potential for a wide range of applications.

\paragraph*{Existing applications} In AR, global illumination methods require a complex initialisation process to provide a good rendering and lack flexibility for new viewpoints. For retexturing applications, JOLIMAS can be used instead of relying on global or local illumination models \cite{morgand2016empirical, morgand2017amultiple}. For scene analysis methods, it is beneficial to have a quick way to retrieve the state of the light sources (turned on and off), which is hard to obtain with state of the art methods compared to JOLIMAS, as shown in \cite{morgand2017geometric}. 

\paragraph*{Potential applications}
In an autonomous driving assistance system (ADAS), distinguishing efficiently specularities from white road lines is important. It can use JOLIMAS as a specularity prediction system to make the specularity detection less ambiguous and the driving safer \cite{kim2021specular}. In Diminished Reality, when removing an object from an image stream, it is difficult to synthesize light cues such as specularities to provide a plausible rendering. \cite{said2017image} proposed an inpainting approach of partially cut specularities but lacks a light source representation in cases where there are not enough data to recover the specularity due to the object removal. In local illumination models, it is difficult to accurately triangulate the light source position without a proper specularity model, as shown in \cite{hadj2019can}. In machine learning, collecting a proper image database with segmented specularities is a long and tedious process which could be tremendously spedup with a proper specularity geometric model for both specularity synthesis, segmentation and removal applications. The current state of the art relies heavily on synthetic image databases as shown in \cite{lin2019deep}, or builds a database from real images using specularity detection system such as \cite{morgand2014generic, kim2021specular}.
However, Dual JOLIMAS is limited to uniformly curved surfaces and fails for changes in the surface curvature.

\subsection{Contributions}

The relationship between surface curvature and specularity movement was studied in \cite{blake1988geometry}, showing promising results for a better understanding of the specularity behaviour from a geometric standpoint. In a context where the camera pose and the scene geometry are known, the specularity movement can thence be predicted. However, the question of explaining and exploiting the specularity's shape transformation due to curvature changes has been left unresolved. In this paper, we present Canonical Dual JOLIMAS, a generalisation of the JOLIMAS model family. It allows for specularity prediction by embedding  the link between curvature and specularity shape to alleviate the geometric assumptions on the surface shape made by the two previous versions of JOLIMAS. The ability to deal with general surfaces widely opens the applicability of JOLIMAS to real world problems.

\section{Background and Problem Statement}
\label{sec:background}

In multiple view geometry, we distinguish several cases to reconstruct a static object from perspective projections. The easiest case is where an object is directly observed. Points, lines or other geometric primitives are modelled in the image by simple perspective projection according to camera pose. 
For an object observed through a perfect mirror, the visual cues are modelled in the image by computing the symmetry according to the normal of the mirror. 
Estimating the shape and extent of a light source observed in an image is an ill-posed problem.
For a point light source directly observed in an image, its intensity combined with the light sensitivity of the camera makes the appearance of the light source ambiguous, even when the camera sensor parameters are fixed. When reflected on a surface, due to material properties, the reflection of the light source will also be ambiguous and will not exactly match the original shape of the light source.
As a consequence, geometrically recovering a light source in a consistent and accurate way is difficult because this light source shape increases drastically according to the global intensity of the image. 
In the presence of a specular surface, in addition to the intensity of the light source and sensor light sensitivity, the image of the point light source is observed through a shiny surface with a mirror-like behaviour and is affected by the surface's material (reflectance and roughness). 
This implies that this image does not match the perspective projection of the light source shape. 
This mismatch between the real shape of the light source and the one observed in the image seems to be, up to scale, the light source real size.
We can link specularity geometric modelling methods to Structure-from-Motion methods for 3D reconstruction from mirror reflections such as \cite{kuthirummal2007flexible, mitsumoto19923, savarese2005local}.
Most of the theory behind the JOLIMAS model relies on the existence of a fixed 3D object in space reflected by a specular surface.  If we find the relationship between surface curvature (mirror or specular) and 3D object to reconstruct (ellipsoid in the case of JOLIMAS),  it would allow us to generalise the specularity prediction model to any curvature of the specular surface.

\begin{figure}[!ht]
\centering
\includegraphics[width=\linewidth]{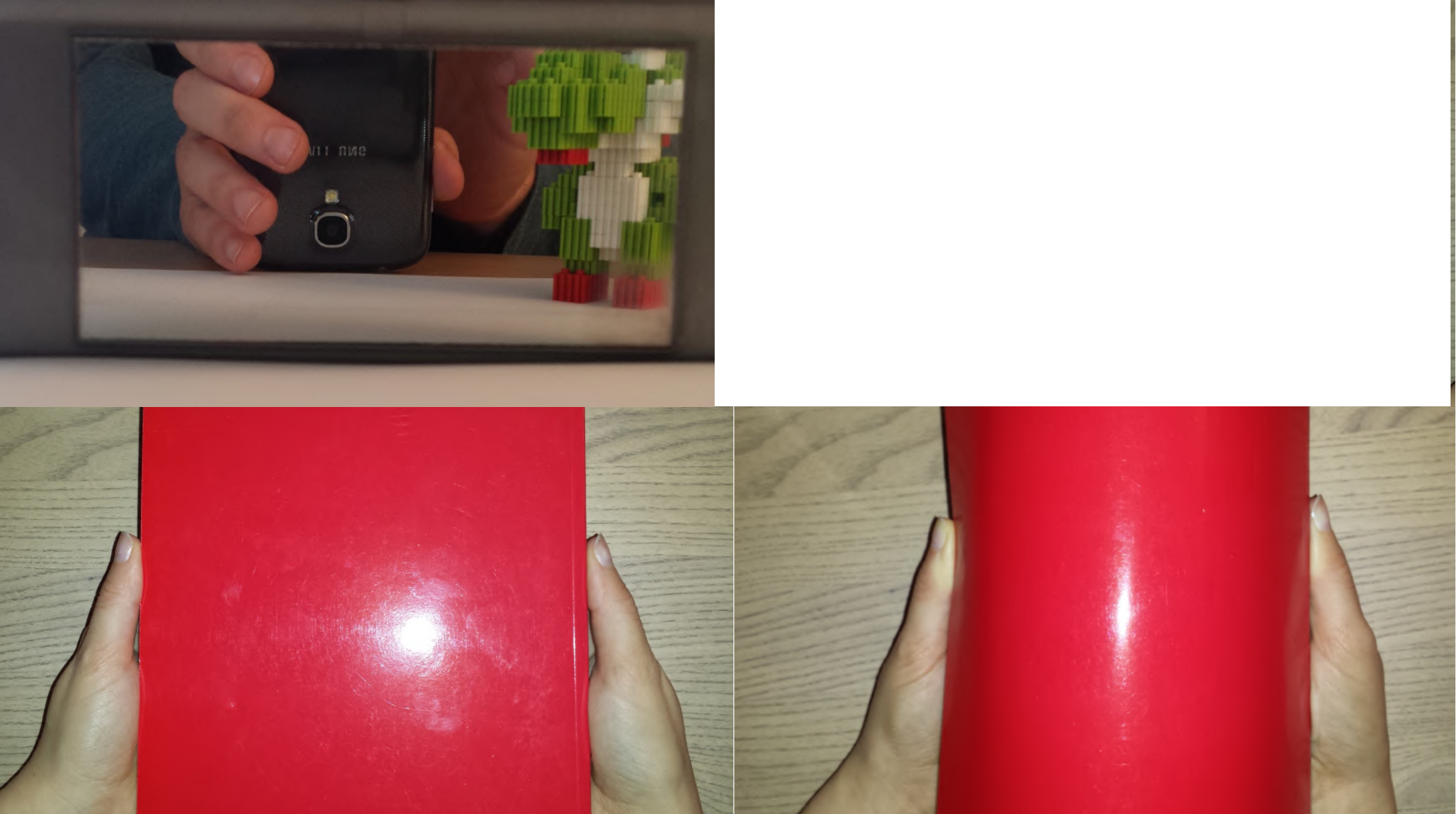}
\caption{Mirrored images of a 3D scene reflected by two types of surfaces. The top row shows two mirror objects: a plane (left) and a sphere (right). The bottom row shows a book with a shiny cover bent from a plane (left) to a cylinder (right). 
The reflected images of 3D objects (a plastic statue and a \textit{smartphone} for the mirror surfaces and a light source for the specular surface) are deformed according to the local curvature of the surface. 
During a specularity prediction process, these deformations must be taken into account.}
\label{fig:mirror}
\end{figure}

\begin{figure}[!ht]
\centering
\includegraphics[width=\linewidth]{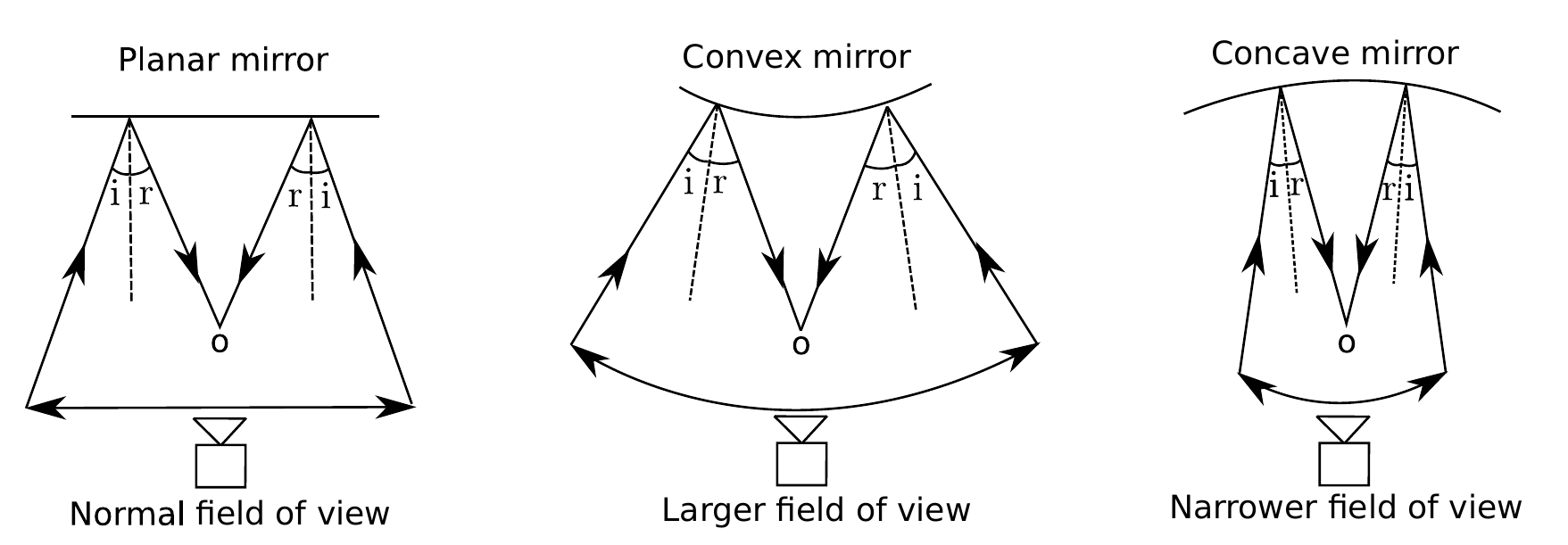}
\caption{Relationship between curvature and field of view in the case of planar, convex and concave mirrors (from left to right). 
The field of view is estimated by computing the reflected rays (r) using the incident rays (i) from the current viewpoint (O). 
We observe that the field of view increases when the mirror bends convex and reduces when the mirror bends concave.}
\label{fig:mirror_fov}
\end{figure}

\begin{figure}[!ht]
\centering
\includegraphics[width=\linewidth]{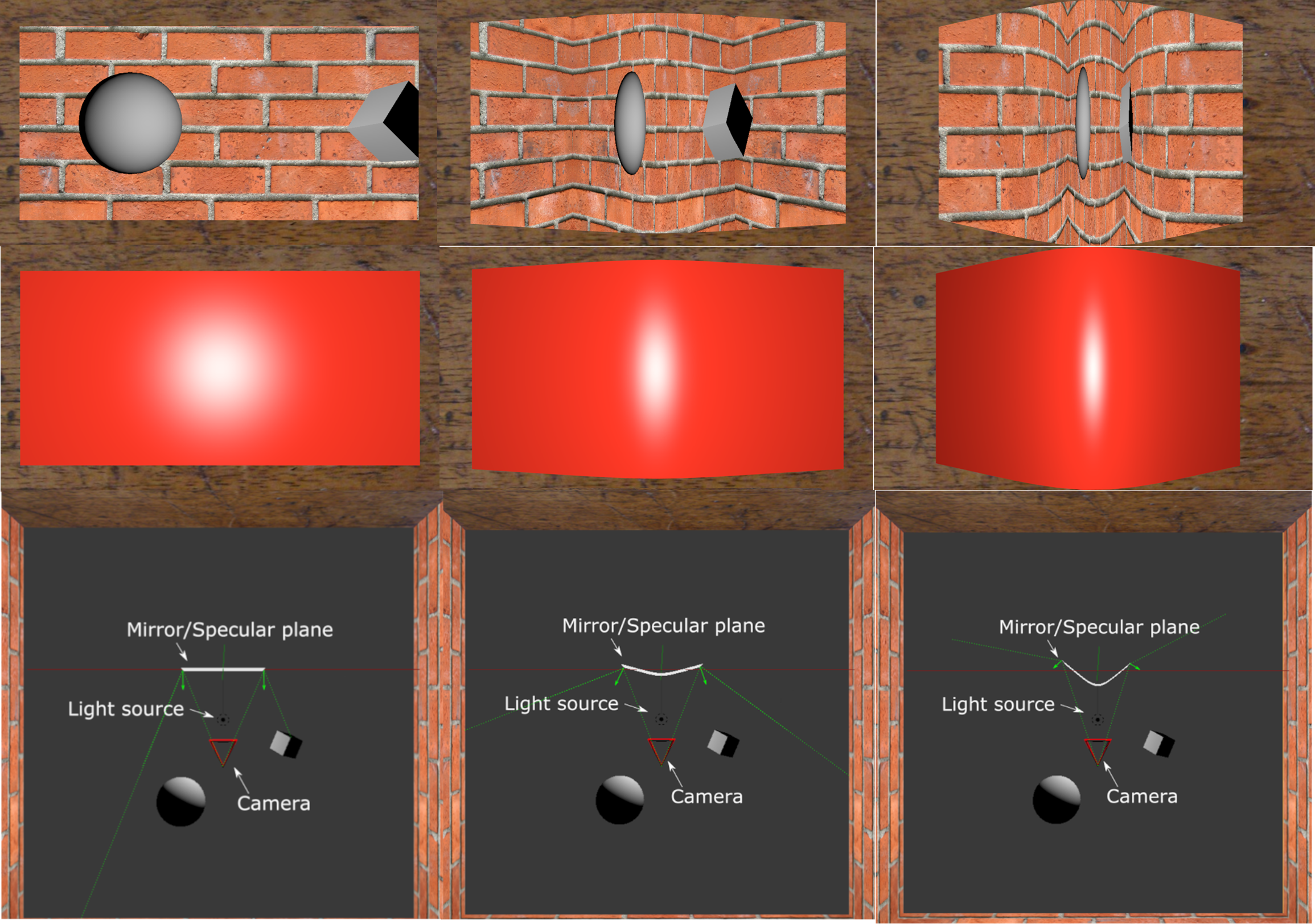}
\caption{Illustration of the relationship between curvature and field of view using a synthetic 3D scene containing a cube, a sphere, a light source and a reflective surface (mirror and plastic surface). 
In order to illustrate the change of curvature of the surface and change of field of view, we curve the reflective surface of the scene progressively and observe changes on the mirror reflections (first row) and of the specularity (second row).
In the third row, an aerial view of the 3D scene is shown and field of view limits are drawn in dotted green lines.
We observe in a similar way to figure \ref{fig:mirror}, that the 3D scene reflected by the mirror is deformed (first row) and that the specularity is also deformed (second row). 
This experiment shows that the field of view increases when the surfaces change from planar to convex when observing mirror-like surfaces. As opposed to the deformation applied to the texture of a surface that we bend which remains at the same physical location, a reflected image is affected by the reflected rays of light of the specular surface.}
\label{fig:synt_mirror}
\end{figure}

Objects observed through a curved mirror are deformed, as illustrated in figure \ref{fig:mirror}.
 The observation is similar for a specularity observed through the red book with a plastic cover. 
 We also observed that in the case of the mirror surface, the field of view seems wider despite the surface area remaining the same. 
 A first hypothesis would be that the curvature change implies field of view variation. 
 Since the surface area remains constant during the deformation, the area of the object image decreases when the surface becomes concave and increases when the surface becomes convex. 
 This link is further highlighted in our synthetic data in figure \ref{fig:synt_mirror}.
 The camera field of view is defined by the observable world that is seen at any given moment and is parameterised by two components: an horizontal and a vertical angles. 
 In the presence of a mirror surface, this field of view can be increased, for instance with a rear view mirror or a catadioptric camera. 
 Indeed, light rays are reflected on the mirror surface, giving a wider field of view. 
 It is possible to change the field of view reflected by this mirror by changing its curvature. 
 More precisely, when one bends a planar mirror to a convex shape, the field of view increases and respectively decreases when bent into a concave shape. This property is illustrated in figure \ref{fig:mirror_fov}.

\section{Proposed Approach - Canonical Dual JOLIMAS}
\label{sec:canonical}
\begin{figure}[!ht]
\centering
\subfigure[]
{
    \includegraphics[width=\linewidth]{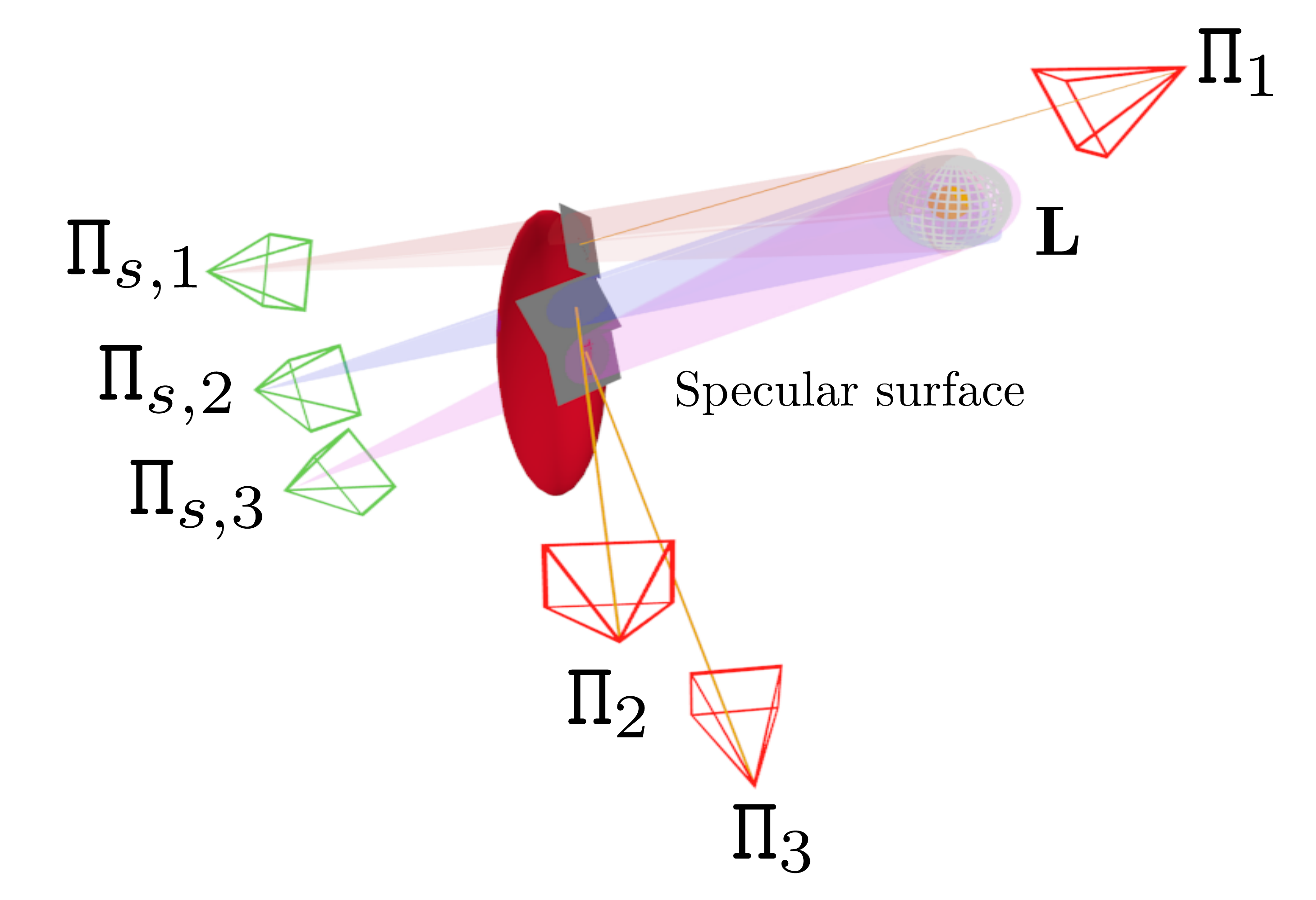}
}
\caption{Illustration of the new canonical dual JOLIMAS model. For each real camera pose ($\Pi_1$, $\Pi_2$ and $\Pi_3$), we create virtual camera poses 
($\Pi_{s, 1}$, $\Pi_{s, 2}$ and $\Pi_{s, 3}$) computed from the brightest point of each specularity. By synthesizing the specularity shape on the tangent plane located at the brightest point for each image, we can reconstruct an ellipsoid located near the light source $\mbf{L}$ in spite of a varying surface curvature.}
\label{fig:canonical}
\end{figure}

We generalise the JOLIMAS model to a canonical form to be able to predict specularities on any surface geometry regardless of its curvature. We first explain the use of the local curvature to make the link between curvature and specularity contour changes.  This canonical dual JOLIMAS is illustrated in figure \ref{fig:canonical}. The remaining assumptions are known camera poses and a known normal map for the surface presenting the specularities.

The concept of an uniformly curved surface refers to the constant curvature that we can find in differential geometry. If, at any point of the surface, the sectional curvature, i.e, the local geometry, is the same, the surface is said to be uniformly curved. For example, a sphere is a surface of constant positive curvature. We will as consider non-uniformly curved any surfaces breaking this condition.

\subsection{Use of Local Curvature}

From the method \cite{healey1988local}, which is an application of \textit{Shape-from-Specularity} consisting in reconstructing the object geometry from the movement of a specularity on a surface, several ideas and hypotheses are interesting to explore in order to find the link between local curvature and specularity shape. Despite the plausible rendering quality that Phong's model provides, \cite{healey1988local} points out that the parameters of Phong's model  are not physics-based, which is unfit for a coherent and realistic specularity modelling. More precisely, \cite{healey1988local} use the Torrance-Sparrow model \cite{torrance1967theory} developed by physicists, giving a detailed formulation on the specular component. This model assumes that a surface is composed of randomly oriented microfacets and having a similar mirror-like behaviour. This model also quantifies facets occluded by other adjacent facets using a geometric attenuation factor. The specular model is then described by three factors:
\begin{equation}
I_S = F D A,
\label{eq:torrance_sparrow}
\end{equation}
with $F$ the Fresnel coefficient, $D$ the function describing the distribution and orientation of a facet and $A$ the geometric attenuation factor.
Coefficient $F$  models the light quantity reflected by each facet. In general, $F$ depends on the incident light ray and a refraction index of the reflecting material. According to \cite{cook1982reflectance}, $F$ defines the specularity colour in order to synthesize realistic images. The distribution function $D$ describes the microfacets orientation according to the surface normal $\vect{N}$. In the case of Torrance-Sparrow\cite{torrance1967theory}, a Gaussian distribution is used to describe $D$ as:
\begin{equation}
D = Ke^{-(\alpha/m)^2},
\label{eq:d_factor}
\end{equation}
where $K$ is a normalisation constant, $m$ a roughness index and $\alpha$ the incident angle. Thus, for a given $\alpha$, $D$ is proportional to the oriented facets to the direction of the vector $\hat{\mbf{H}}$ describing the half-way vector \cite{blinn1977models}.
In order to analyse the specularity shape, it is not necessary to have all of equation \eqref{eq:torrance_sparrow} parameters. $F$ is a nearly constant function of the incident angle for the class of materials with a large extinction coefficient like metals and other strongly reflective materials. For simplicity, \cite{healey1988local} considers $F$ as constant with respect to the viewing geometry and is valid for most reflective materials. Moreover, \cite{healey1988local} shows that the exponential factor in equation \eqref{eq:d_factor} varies rapidly compared to the other terms, allowing one to simplify equation \eqref{eq:torrance_sparrow} to:
\begin{equation}
I_S = K'e^{-(\alpha/m)^2},
\label{eq:torrance_sparrow2}
\end{equation}
where $K'$ is constant.
\cite{healey1988local} also mentions the existence of a brightest point such that the intensity is maximal at this point. By studying the variation of intensity from this point, it is possible to deduce the surface curvature. The link between intensity $I_S$ at a point $\vect{P}$ and the incident angle is given by:
\begin{equation}
| \alpha | = m \sqrt{-\mathrm{log}\frac{I_S}{K'}}.
\end{equation}
The link between the surface curvature and incident angle is given by:
\begin{equation}
\kappa = \frac{\mathrm{d} \alpha}{\mathrm{d} s}\Bigr|_{\substack{\mbf{P_B}}},
\label{eq:curvature_link}
\end{equation}
with $\kappa$ the local curvature for a given direction, $\mathrm{d} \alpha$ a small angle change of the reflected light source and $\mathrm{d} s$ the arc length on the surface at the brightest point $\mbf{P_B}$.  In practice, equation \eqref{eq:curvature_link} is difficult to compute for non-parametric surfaces such as wireframe models. Computing the arc length implies computing a geodesic distance, which is computationally expensive and inaccurate for polygonal or rough surfaces, which often occurs in wireframes. For simplicity, we define angle $\alpha$ as $\alpha = \text{cos}^{-1}(\hat{\mbf{N}} \cdot \hat{\mbf{H}})$ with $\hat{\mbf{N}}$ the normal and $\hat{\mbf{H}}$ the half-way vector $\hat{\mbf{H}} = \frac{\hat{\mbf{L}} + \hat{\mbf{V}}}{\| \hat{\mbf{L}} + \hat{\mbf{V}} \|}$ between the light ray $\hat{\mbf{L}}$ and the observed ray $\hat{\mbf{V}}$.

\subsection{Practical Use on a CAD Model - Limit Angles}
In order to optimise local curvature computation, we propose to use the concept of limit angles. These correspond to angle values $\alpha_{max}^i$ such that for an angle larger than this limit, the specularity intensity is considered null. Limit angles $\alpha_{max}^i$,  
with $i \in \llbracket0, n\rrbracket$, are computed from the specularity contours and particularly the normals $\hat{\mbf{N}}$ associated to the contour points and half-way vectors $\hat{\mbf{H}}$. According to equation \eqref{eq:torrance_sparrow2}, the intensity of the specularity is influenced predominantly by the angle $\alpha$. The  intensity associated to the specularity contour points remains approximately independent of the curvature of the surface, implying that the limit angles keep their values in spite of local curvature changes at the contour points of the specularity.

In our canonical representation, in order to transform the current ellipse associated to the specularity, we compute the brightest point $\mbf{P_B}$ in a similar way as \cite{morgand2017multiple} and its associated tangent plane $T_{\mbf{P_B}}(S)$. We sample $n$ vectors $v_i \in T_{\mbf{P_B}}(S)$ starting from the point $\mbf{P_B}$ within a range of $[0, 2\pi[$.

The choice of $n$ depends on the specularity size and its shape. In practice, we fixed a value of 36 for the presented sequences in figures \ref{fig:results_synt}, \ref{fig:results} and \ref{fig:data_kinect}. This value empirically gives us the best results. When we reach the external contours of the specularity, we compute a limit angle $\alpha_{max}^i$ using $\hat{\mbf{H}}$ and $\hat{\mbf{N}}$ at the orthogonally projected point on the surface $S$. Then, we follow the same vector $v_i$ on $T_{\mbf{P_B}}(S)$ until reaching a point where the angle value is at $\alpha_{max}^i$, which is the limit associated to the vector. The transformed ellipse is computed from the new computed contour points.
\begin{figure}[!ht]
\centering
\subfigure[]
{
    \includegraphics[width=\linewidth]{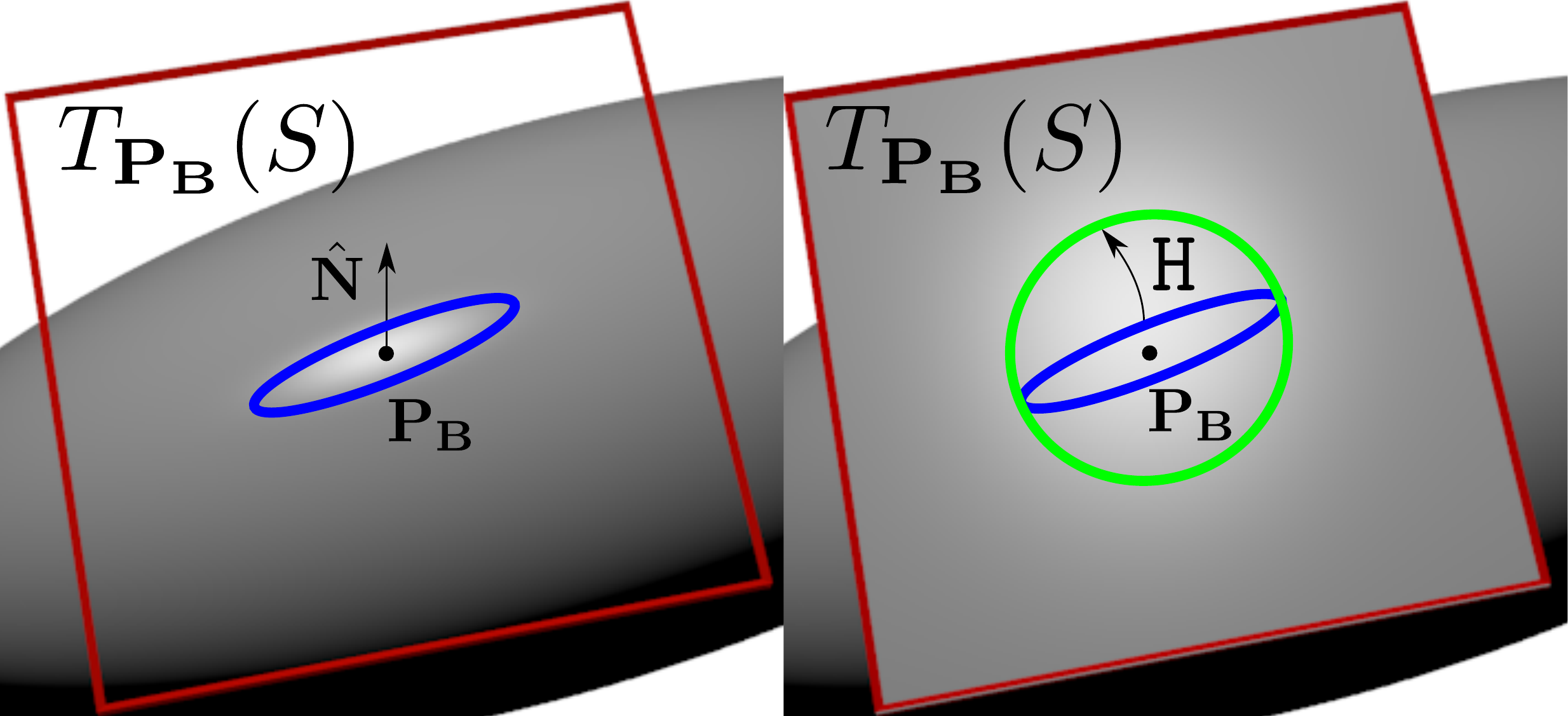}
}
\subfigure[]
{
    \includegraphics[width=\linewidth]{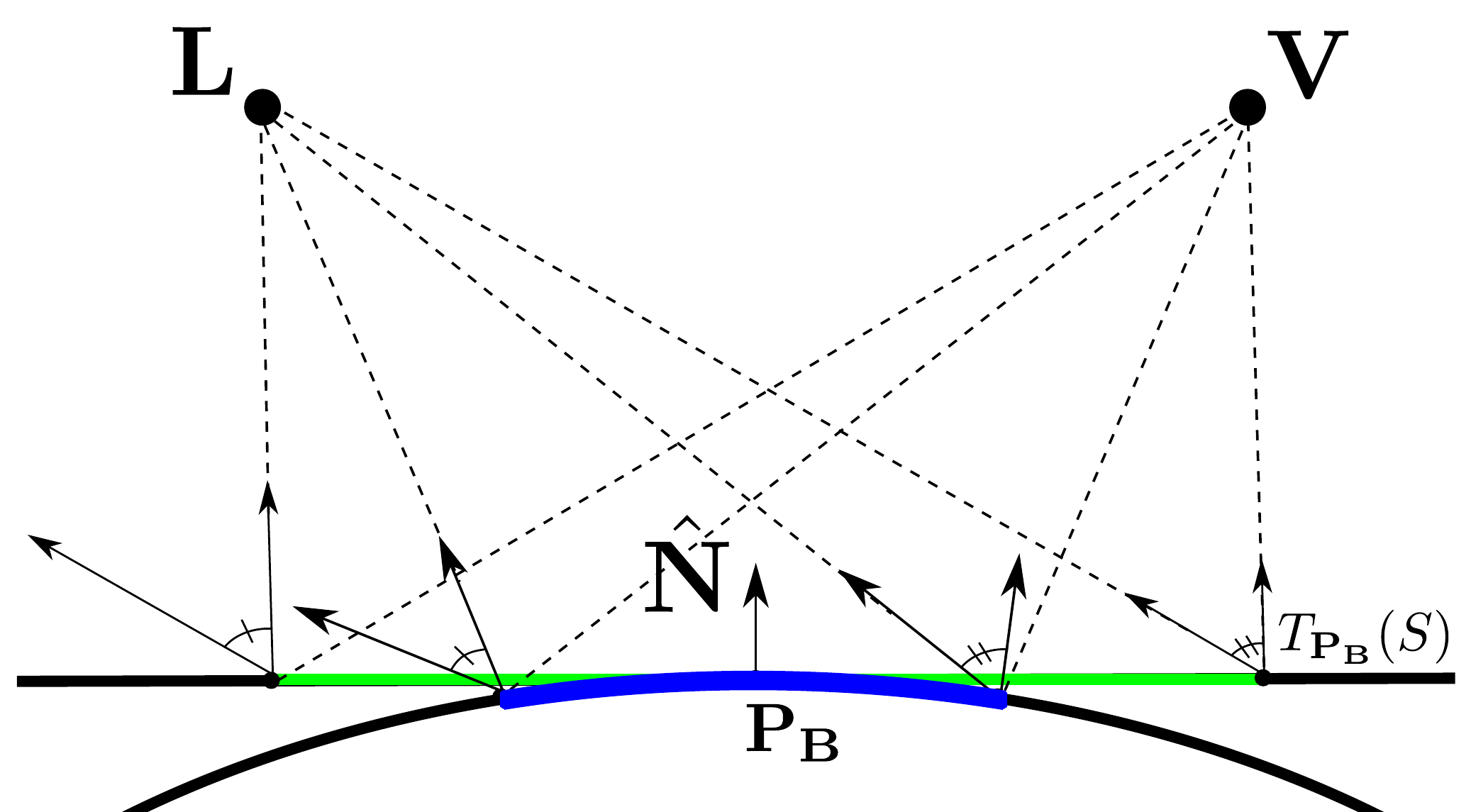}
}
\caption{Ellipse transformation from its shape on the curved surface $S$ to the canonical representation on the tangent plane $T_\mathbf{P_B}(S)$ at the brightest point $\mbf{P_B}$. (a) shows the transformation process starting from the ellipse fitted to the specularity contours (blue). We apply this transformation in order to obtained the corrected ellipse (green) on the tangent plane $T_\mathbf{P_B}(S)$. In (b), we show that the limit angles keep their values at the specularity contours without being affected by the surface curvature (blue for the curved surface and green for the plane).}
\label{fig:warping}
\end{figure}
From the transformed ellipses, we are able to use the same ellipsoid reconstruction formalism using virtual cameras, as in dual JOLIMAS \cite{morgand2017multiple}.

\subsection{Specularity Prediction from Limit Angles}
During the specularity prediction step, a similar process to  \cite{morgand2017multiple} is used to predict the shape of the specularity for a new viewpoint. We compute the brightest point for the new viewpoint and we project the reconstructed ellipsoid which gives us the specularity shape on the tangent plane $T_{\mbf{P_B}}(S)$. We then compute the limit angles on $T_{\mbf{P_B}}(S)$ and move the specularity contours on $S$ until reaching the limit angle on $T_{\mbf{P_B}}(S)$ by taking into account the local curvature of the surface S. 

\section{Experimental Results}
\label{sec:results}
We present quantitative results on synthetic data to assess the quality of our specularity prediction and qualitative results on real data compared with the state of the art method\cite{morgand2017multiple}. We evaluate the quality of our specularity prediction ability by comparing the 2D distance between specularity prediction and the contours of the
detected specularity.

\subsection{Synthetic Data}
\label{sec:validation_synt}

\begin{figure}[!ht]
\centering
\subfigure[]
{
    \includegraphics[width=\linewidth]{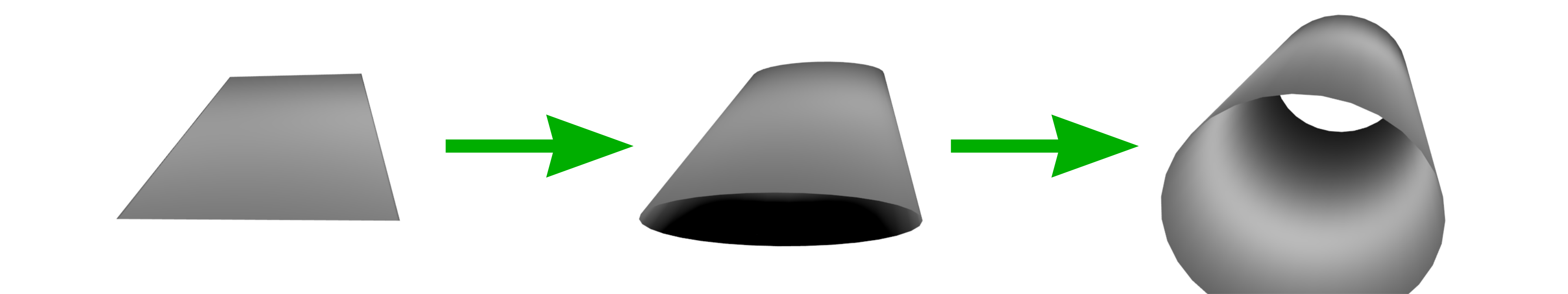}
    \label{fig:curvature_seq}
}
\subfigure[]
{
    \includegraphics[width=\linewidth]{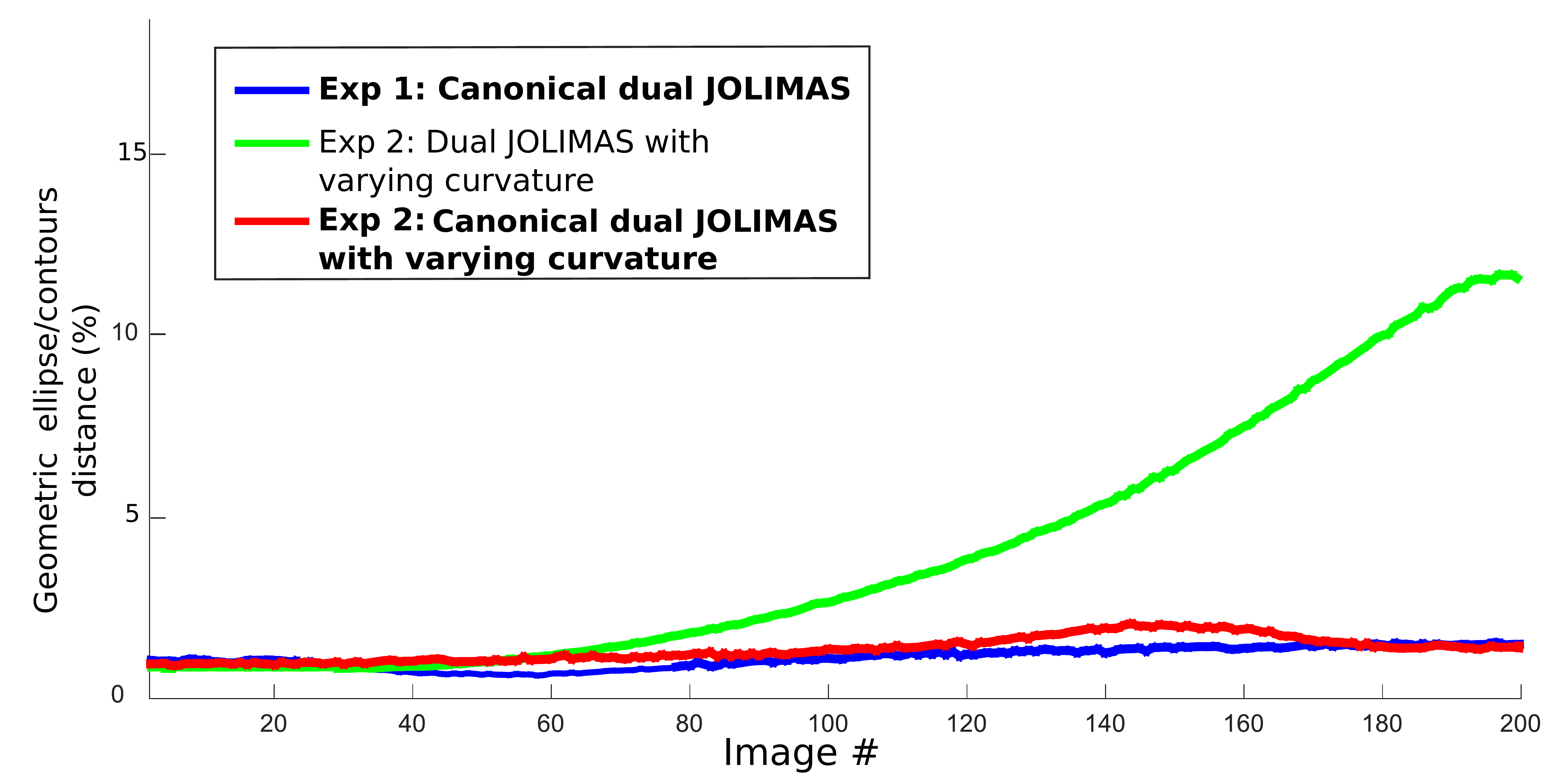}
    \label{fig:synt_cylinder}
}
\caption{Synthetic experiment to evaluate the accuracy of specularity prediction when curvature changes. (a) illustrates how we vary the curvature from a plane to a cylinder. In this test, our ellipsoid is reconstructed from six images from a planar surface. \textbf{Exp 1} reconstructs the ellipsoid (the canonical dual JOLIMAS model) for each curvature changes while \textbf{Exp 2} reconstructs it only once from the first 6 images. The prediction is conducted from this reconstruction for every future viewpoint where the plane is morphed into a cylinder. We observe that canonical JOLIMAS through \textbf{Exp 1} (in blue) and \textbf{Exp 2} (in red) has a low error in terms of specularity prediction as opposed to dual JOLIMAS which has an increasing error following the curvature change in \textbf{Exp 2} (in green).}
\label{fig:synt_cylinder_main}
\end{figure}

\paragraph*{Plane/cylinder sequence} 
In order to test the improvement of our canonical dual model compared to the dual one, we reuse the experiment from \cite{morgand2017multiple}. It consists of a video sequence composed by 300 synthetic images of a cylinder varying in radius as shown in figure \ref{fig:curvature_seq}, in order to simulate the curvature change from a plane to a cylinder. The goal is to assess the specularity shape prediction on a varying curvature by computing the 2D distance between ellipses given by the predicted specularity from the compared JOLIMAS methods and the fitted ellipse from detected specularity contours. Each curvature change is observed through 6 different synthetic viewpoints.  This test is composed of two experiments. \textbf{Exp 1} reconstructs the ellipsoid (canonical JOLIMAS model) for each curvature changes while \textbf{Exp 2} reconstructs it only once from the first 6 images. The prediction is done from this reconstruction for every future viewpoints where the plane is morphed into a cylinder.
We present the results in figure \ref{fig:synt_cylinder}. For \textbf{Exp 1}, our model fits almost perfectly the specularity contours with an average error of 1\%. For \textbf{Exp 2}, we observe that our canonical model can adapt to curvature changes of the surface throughout the sequence, while keeping an optimal precision of the ellipse estimation, the predicted specularity, with an average error of 2\%. The previous dual model has an increasing error following the curvature change, reaching up to 12\% with an average error of 6\%.

\begin{figure}[!ht]
\centering
\subfigure[]
{
    \includegraphics[width=0.287\linewidth]{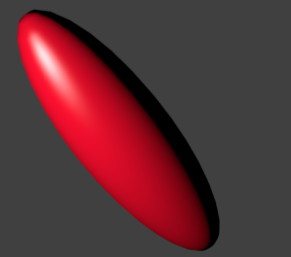}
}
\subfigure[]
{
    \includegraphics[width=0.57\linewidth]{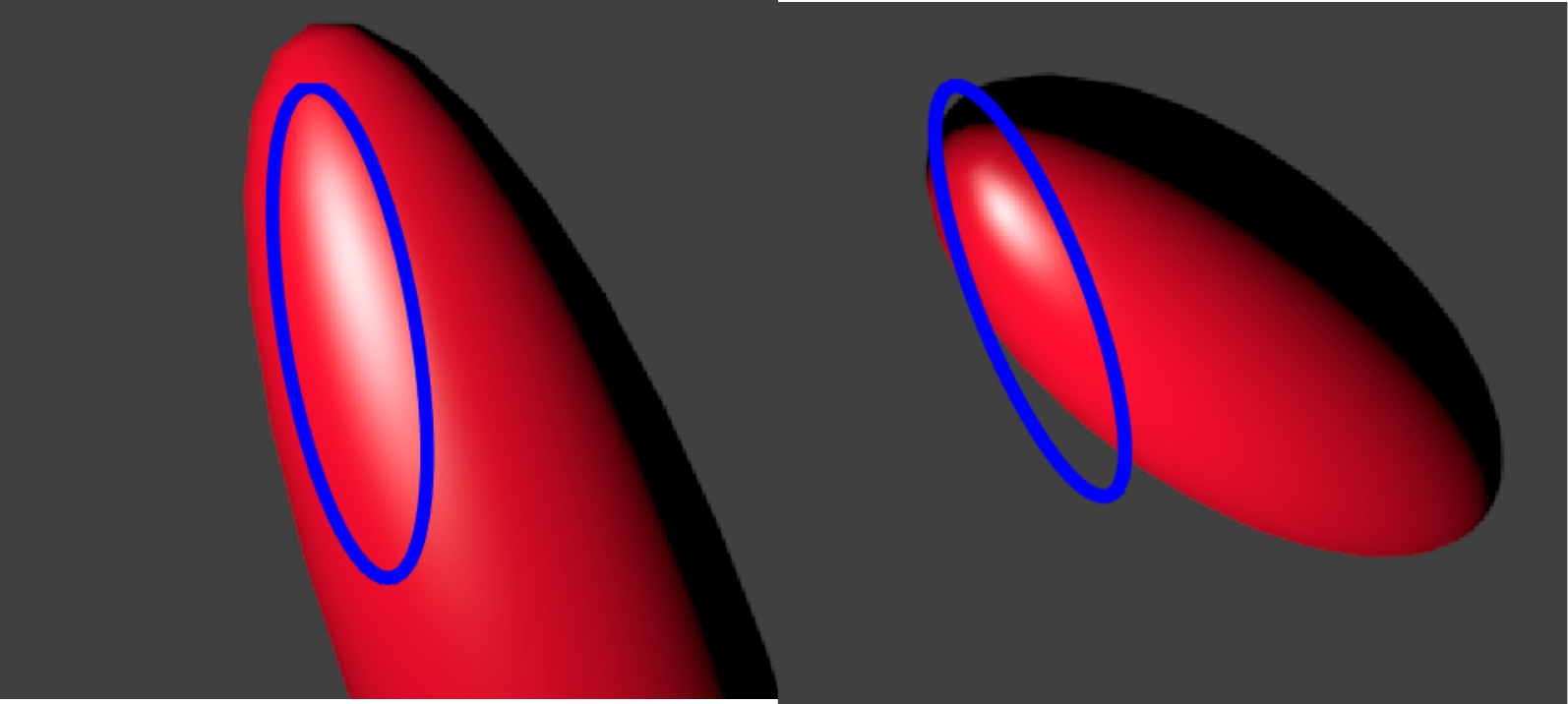}
}
\subfigure[]
{
    \includegraphics[width=0.59\linewidth]{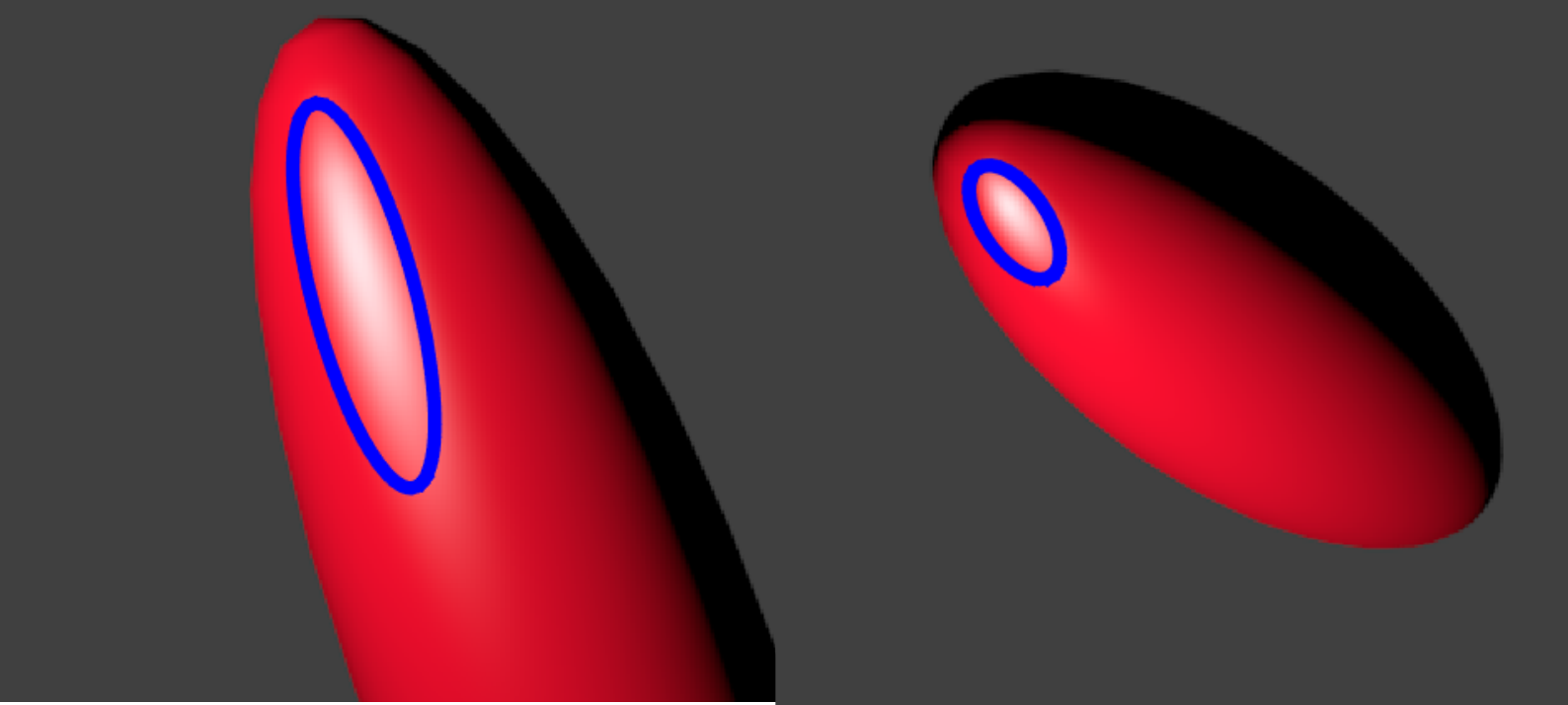}
}

\caption{Specularity prediction results for our ellipsoid synthetic sequence. In (a), we show that the ellipsoid object is not uniform in curvature. In (b) and (c), we show an image pair of the sequence and the specularity prediction results (blue ellipses) of canonical dual JOLIMAS (b) and canonical JOLIMAS (c). The image on the left corresponds to the curvature used to reconstruct the model and the one on the right illustrates prediction for stronger curvature. We see that our canonical JOLIMAS gives better results than dual JOLIMAS for both curvatures, as dual JOLIMAS performs almost well on the first image but badly on the varying curvature on the right image.}
\label{fig:results_synt}
\end{figure}

\paragraph*{Ellipsoid sequence}
In order to check the validity of the canonical JOLIMAS model on a simple example, we created a synthetic sequence of 80 images containing an ellipsoid object which is a shape with non-uniform curvature. Using the first 6 frames, we reconstruct our canonical dual JOLIMAS model and the dual JOLIMAS model\cite{morgand2017multiple}. Then we use these reconstructions to perform specularity prediction on the remaining frames of the sequence.
These results are illustrated qualitatively in figure \ref{fig:results_synt}. We can see that our specularity prediction of canonical dual JOLIMAS outperforms dual JOLIMAS in terms of position and shape prediction of the ellipse. In this sequence, our method achieves an average error for specularity prediction of 2.4\% while \cite{morgand2017multiple}'s error increases rapidly with curvature change, with an average error of 20.2\%.

\subsection{Real Data}
\label{sec:real_data}
We present specularity prediction on two real sequences containing a metal rocket replica and a red book with a shiny cover. The reconstruction of canonical dual JOLIMAS and dual JOLIMAS is done from the same six images for each sequence. The specularity prediction is done for the remaining images of the sequences following the ones used for reconstruction.

\begin{figure}[!ht]
\centering
\subfigure[]
{
    \includegraphics[width=0.9\linewidth]{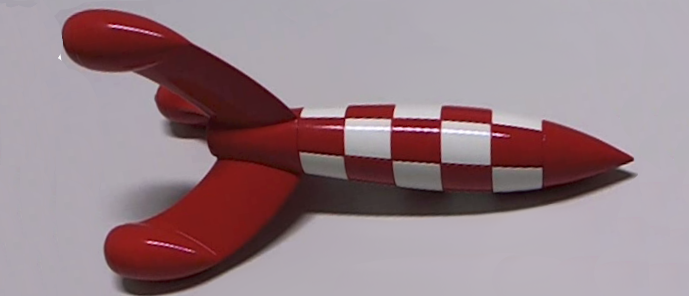}
}
\subfigure[]
{
    \includegraphics[width=0.9\linewidth]{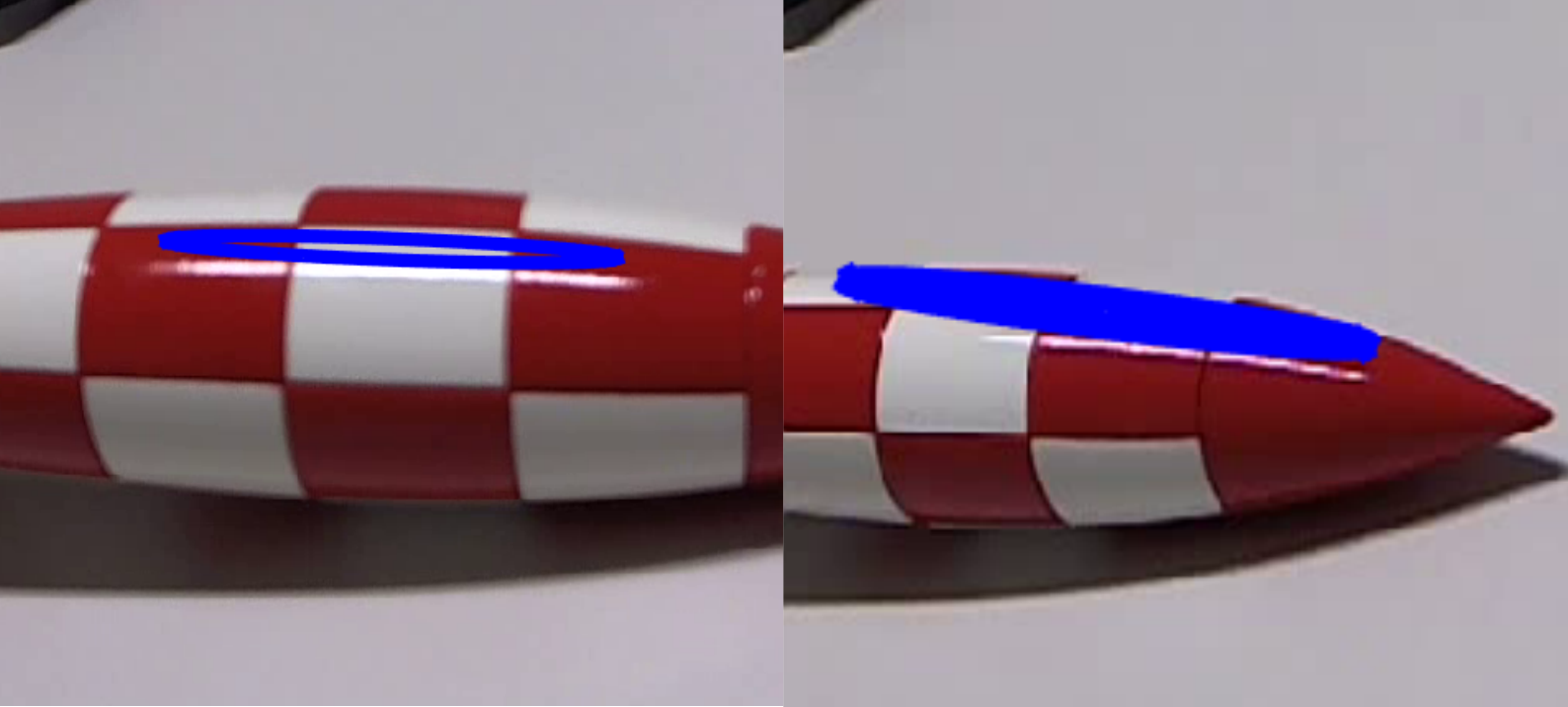}
}
\subfigure[]
{
    \includegraphics[width=0.9\linewidth]{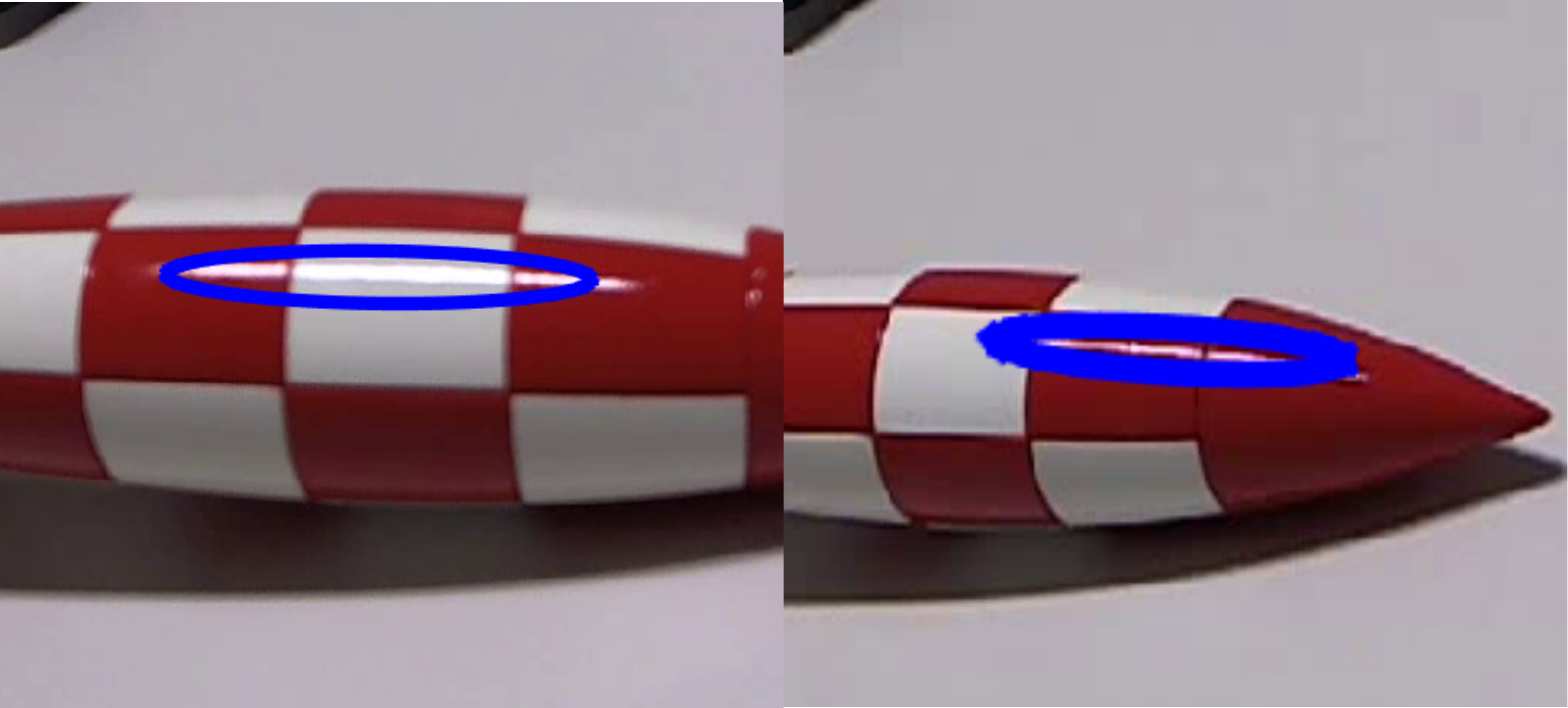}
}

\caption{Specularity predictions on the real sequence presented in \cite{morgand2017multiple}. (a) shows the object of interest: a real rocket replica. We present a pair of images for each model showing the prediction results (blue ellipses) of dual JOLIMAS (b) and canonical dual JOLIMAS (c). Our new approach models the position and shape of the specularities in a more accurate way compared to dual JOLIMAS, with an average error of 1.3\% compared to 2.1\% with dual JOLIMAS, by comparing the 2D distance between specularity prediction and the contours of the
detected specularity.}
\label{fig:results}
\end{figure}

\paragraph*{Rocket replica sequence}
We introduced this 1410 frames sequence in \cite{morgand2017multiple} and although the results were already acceptable with an error on average of 2.1\%, the central part of the rocket replica does not have a constant curvature, which negatively affected the specularity prediction accuracy of dual JOLIMAS. We observe in figure \ref{fig:results} that canonical dual JOLIMAS outperforms  dual JOLIMAS in terms of specularity prediction with an average error of 1.3\%.

\paragraph*{Kinect sequence}
To show the potential of our method in a more complex context, we took a sequence with 146 real images from a Kinect v2\footnote{https://developer.microsoft.com/en-us/windows/kinect}, where depth information is computed from a Time of Flight (ToF) system.
Kinect v2 is mainly supported on Windows 8 and we had to use the Libfreenect2 library from the OpenKinect\footnote{https://github.com/OpenKinect/libfreenect2} project. This library allows one to retrieve infrared information as a  $512\times424$ image and colours as a  $1920\times1080$ image from the sensor. This library also provides tools to synchronise data to obtain a $512\times424$ depth image.
In order to reconstruct our canonical dual JOLIMAS, it is necessary to retrieve surface normal information where the specularities appear. In the case of Kinect v2, a normal map needs to be deduced from the point cloud given by the depth image. However, Kinect v2 does not provide depth for objects reflecting infrared rays such as mirrors and specular surfaces. In addition, it has holes in occluded areas in the infrared image due to the baseline between the RGB and infrared cameras. These areas correspond in general to occluding contours of objects.
To address these issues, we first applied an inpainting algorithm \cite{telea2004image} on the depth image in order to fill the potential holes. For larger areas, we could also use methods such as \cite{chican2014constrained, bousefsaf2018image}. In our sequence, these holes are generally numerous but small, which allows us to interpolate neighbouring information in the image. 

Once the depth image is inpainted, we use the Point Cloud Library (PCL)\footnote{http://pointclouds.org} to compute the normal map. This map is obtained using 3D points within a radius around each point depending on the point cloud density. The optimal search radius used here in our experiments is 3 cm. 

\begin{figure}[!ht]
\subfigure[]
{
    \includegraphics[width=\linewidth]{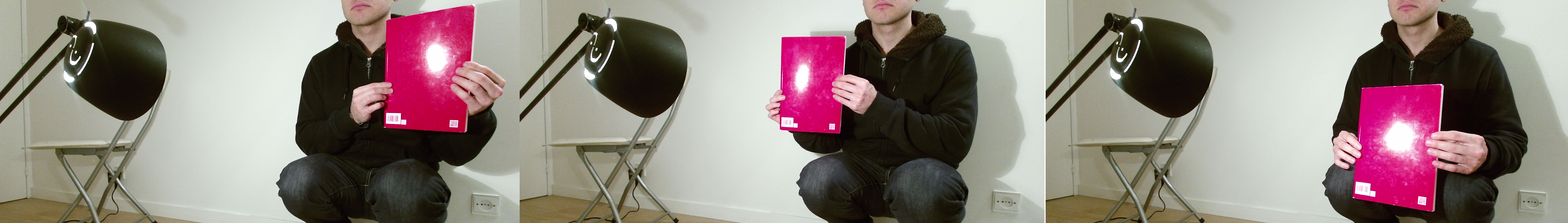}
}
\subfigure[]
{    
    \includegraphics[width=\linewidth]{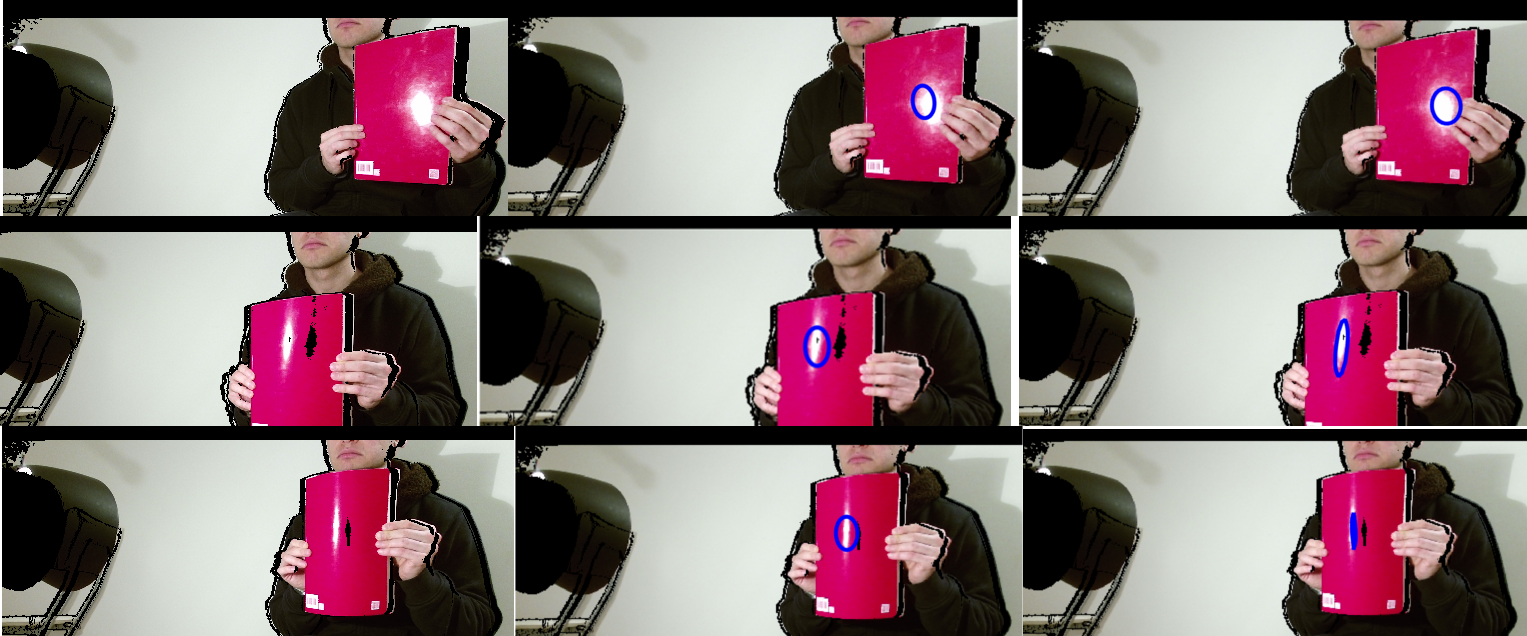}
}

\caption{Images used for model reconstruction (a) and results of specularity prediction on the Kinect sequence for three images showing different curvatures (for each row) (b). Images on the first column are the source images, the middles ones correspond to dual JOLIMAS specularity prediction and the ones on the right are from canonical dual JOLIMAS.}
\label{fig:data_kinect}
\end{figure}

We present in figure \ref{fig:data_kinect} our specularity prediction results for the Kinect sequence for both canonical dual and dual JOLIMAS. As expected, dual JOLIMAS is not able to handle the curvature change of the book, as opposed to canonical dual JOLIMAS. Note that even with a limited $100\times150$ resolution of the book surface in the image and local noise in the normal map, we manage to produce qualitatively correct results.

\section{Limitations}

\subsection{Image Resolution, Field of View and Brightest Point Estimation}
Our JOLIMAS model is sensitive to the brightest point estimation. We highlighted the importance of computing an accurate one per specularity in previous work\cite{morgand2016empirical, morgand2017geometric, morgand2017multiple, morgand2017amultiple}. As we show in figure \ref{fig:synt_mirror}, curvature changes imply a variation in the field of view seen by the reflected surface. However, it means that for a similar amount of pixels in the image, we can see more or less information of the scene. Even though we have a subpixel estimation of the brightest point, it can be harder to estimate depending on the surface curvature and the image resolution.
\subsection{Sensitivity to Depth Map Noise}
Our canonical dual JOLIMAS model proposes a new solution to the specularity prediction problem for non-uniformly curved objects (ellipsoid, plastic rocket replica) as well as objects with dynamically changing curvature (book bent in real-time throughout the sequence). Our experiments showed the critical relationship between 3D reconstruction quality (from the CAD model and Kinect v2) and stability/accuracy of the prediction. In the case of Kinect, it is quite difficult to obtain a prediction without jittering caused by the normal map due to the raw noise of the sensor depth data. Moreover, due to our method being  sensitive to the quality of camera pose and normal estimate of the surface, the prediction ability of the specularity shape can vary according to the size of the detected specularity in the image and the image size in general.
\subsection{Perfectly Specular Surfaces and Non-Elliptical Shaped Specularity}
In the case of perfectly specular surfaces like a mirror and depending on the camera exposure, it is possible to predict a specularity matching closely the light source shape like a light bulb or fluorescent light source. A mismatch can however happen when the surface has been scratched irregularly, altering the specularity shape greatly. As a consequence, the observed specularity shape could break our elliptical shape specularity assumption. This issue could be lifted by doing a full 3D reconstruction of the reflected light source from the specularity shape and dropping our ellipsoid reconstruction. However, as shown in JOLIMAS in most cases, the specularity shape will follow an ellipse shape which will be enough for most AR applications. Moreover, 3D reconstruction from only specularity contours is highly ambiguous and is closer to space carving than to keypoint matching and triangulation.

\section{Future Work}
In terms of applications, improvements in the specularity prediction capability of canonical dual JOLIMAS will naturally improve the already presented applications such as retexturing \cite{morgand2017multiple}. It will also provide a better light source position for local illumination methods to improve the rendering quality of AR applications, since canonical dual JOLIMAS relies on the Torrance-Sparrow local illumination model\cite{torrance1967theory}. Note that the geometric constraints of the JOLIMAS model are much more important than the illumination model choice to obtain an optimal ellipsoid reconstruction. It would be interesting to show a larger comparison of the reflectance model used and its impact on the rendering quality of AR applications from Phong, Torrance-Sparrow to GGX\cite{walter2007microfacet}. After reconstructing the specular term from the JOLIMAS model, an interesting step would be to jointly reconstruct the diffuse part through a larger optimisation problem. The additional information would benefit camera pose estimation, material estimation methods  but also depth and normal maps refinement as the reconstructed illumination model would refine them. It could also be interesting to study the impact of different materials (roughness and reflectance) on the specularity shape and also on the specularity prediction accuracy. This study could potentially lead to genericity improvements of the model. Using RGB-D data, it would be interesting to design an algorithm to ensure temporal and spatial coherence, allowing one to offer an optimal stability even in difficult cases. A curvature interpolation of the surface and of the specularity shape change between two images could be interesting to test. However, RGB-D data are not mandatory, since one can use constrained SLAM such as \cite{tamaazousti2016constrained}, providing the camera pose and the geometry of the object of interest, as shown with the rocket replica in section \ref{sec:real_data}. In cases where the highlight looks similar to the light source, a possible solution would be to use Structure-from-Motion to reconstruct the non-elliptic specularity shape.
Another interesting application would be to use JOLIMAS specularity prediction ability as a labeling helper tool for deep learning processes, such as specularity detection and light source modelling.

\section{Conclusion}
We have presented a canonical representation of the previous dual JOLIMAS model \cite{morgand2017multiple}, where we have addressed the main limitation of the previous approach: the hypothesis of a fixed and constant surface curvature. Our method generalises the JOLIMAS model to any surface geometry while improving the quality of specularity prediction without compromising performance. We highlighted the link between surface curvature and specularity shape provided by \cite{healey1988local}. By using the physics-based Torrance-Sparrow local illumination for modelling the specularity shape as the light interaction with the material, we computed limit angles that are angle values at the contours of the specularity being constant for any curvature. This has allowed us to deform the specularity to fit the JOLIMAS reconstruction \cite{morgand2017multiple} and adapt the shape of the predicted specularity for new viewpoints. Specularity prediction using our new model was tested against \cite{morgand2017multiple} on both synthetic and real sequences with objects of varying shape curvatures given by the CAD model or using a Kinect v2 depth camera. Our method outperforms previous approaches for the specularity prediction task in real-time. 

\bibliographystyle{plain}
\bibliography{egbib}

\end{document}